\documentclass{article}

\usepackage[nonatbib,preprint]{neurips_2025}
\usepackage[numbers]{natbib}

\newcommand{\m}[1]{\mathbf{#1}}

\usepackage{amsmath,amsfonts,bm}

\def\eqref#1{equation~\ref{#1}}
\def\1{\bm{1}}

\def\vr{{\bm{r}}}

\def\vx{{\bm{x}}}

\def\mX{{\bm{X}}}

\DeclareMathAlphabet{\mathsfit}{\encodingdefault}{\sfdefault}{m}{sl}
\SetMathAlphabet{\mathsfit}{bold}{\encodingdefault}{\sfdefault}{bx}{n}

\usepackage[utf8]{inputenc}
\usepackage[T1]{fontenc} 
\usepackage{url}            
\usepackage{booktabs}      
\usepackage{amsfonts}      
\usepackage{nicefrac}     
\usepackage{microtype}    
\usepackage[dvipsnames]{xcolor}      
\usepackage{array}
\usepackage{graphicx}
\usepackage{amsmath}
\usepackage{multirow}
\usepackage{booktabs} 
\usepackage{array}  
\usepackage{hhline}
\usepackage{cellspace}
\usepackage{caption}
\usepackage[colorlinks=true, urlcolor=blue]{hyperref}
\usepackage{footmisc}
\newcolumntype{C}[1]{>{\centering\arraybackslash}p{#1}}

\title{Mixture of Sparse Attention: Content-Based Learnable Sparse Attention via Expert-Choice Routing}

\author{
  Piotr Piękos$^{1}$, Róbert Csordás$^{2}$, Jürgen Schmidhuber$^{1}$\\[1ex]
  $^{1}$KAUST, AI Initiative, Thuwal, Saudi Arabia \\
  $^{2}$Stanford University, Stanford, CA, USA\\[1ex]
  \texttt{piotr.piekos@kaust.edu.sa} \\ \texttt{rcsordas@stanford.edu} \\ \texttt{juergen.schmidhuber@kaust.edu.sa}
}

\begin{document}

\maketitle

\begin{abstract}
Recent advances in large language models highlighted the excessive quadratic cost of self-attention.
Despite the significant research efforts, subquadratic attention methods still suffer from inferior performance in practice. 
We hypothesize that dynamic, learned content-based sparsity can lead to more efficient attention mechanisms.
We present Mixture of Sparse Attention (MoSA), a novel approach inspired by Mixture of Experts (MoE) with expert choice routing. MoSA dynamically selects tokens for each attention head, allowing arbitrary sparse attention patterns.
By selecting $k$ tokens from a sequence of length $T$, MoSA reduces the computational complexity of each attention head from $O(T^2)$ to $O(k^2+T)$. This enables using more heads within the same computational budget, allowing higher specialization. We show that among the tested sparse attention variants, MoSA is the only one that can outperform the dense baseline, sometimes with up to 27\% better perplexity for an identical compute budget. 
MoSA can also reduce the resource usage compared to dense self-attention. 
Despite using torch implementation without an optimized kernel, perplexity-matched MoSA models are simultaneously faster in wall-clock time, require less memory for training, and drastically reduce the size of the KV-cache compared to the dense transformer baselines.
\end{abstract}

\begin{center}
  \href{https://github.com/piotrpiekos/MoSA}{\textcolor{blue}{\texttt{https://github.com/piotrpiekos/MoSA}}}
\end{center}

\section{Introduction}

Modern transformer architectures \cite{vaswani2017attention} have proven to be highly effective for sequence modeling tasks and are the key to the success of large language models (LLMs; \cite{brown2020language, touvron2023llama, team2024gemini, grattafiori2024llama}). One of the key components of their success is the attention mechanism, which enables dynamic information propagation by computing weighted sums of past states when processing each token. This results in high computational and memory complexity, both quadratic in sequence length. The key to the success of LLMs is the ever-increasing model sizes and context windows. Training and deploying these models becomes increasingly prohibitive. Furthermore, the KV-cache memory footprint during inference presents a significant bottleneck, limiting practical deployment scenarios and increasing operational costs.

This led the researchers to explore alternative approaches. State Space Models~\cite{gu2020hippo, gu2021efficiently, gu2023mamba, wang2024state, yang2025gated} capture long‐range dependencies with just a handful of state variables rather than relying on full attention matrices. They, however, fall short of full self-attention in terms of practical performance. To counteract lossy compression of State Space Models, a recent line of work investigates hybrids that combine quadratic attention and linearized memories~\cite{park2024mambaformer, zuo2022efficient, lieber2024jamba}.  Linear attention~\citep{katharopoulos2020transformers,schlag2021linear,schmidhuber1992learning}\footnote{Note that {\em unnormalized linear transformers} (with "linear attention") were first published in 1992 under the name {\em fast weight controllers}~\cite{schmidhuber1992learning} or {\em fast weight programmers}.} optimizes the attention cost by changing the order of the operations in the attention after removing nonlinearity. However, it also performs poorly compared to quadratic attention~\citep{qin2022-devil}.

\begin{figure}
    \centering
    \includegraphics[width=1\textwidth]{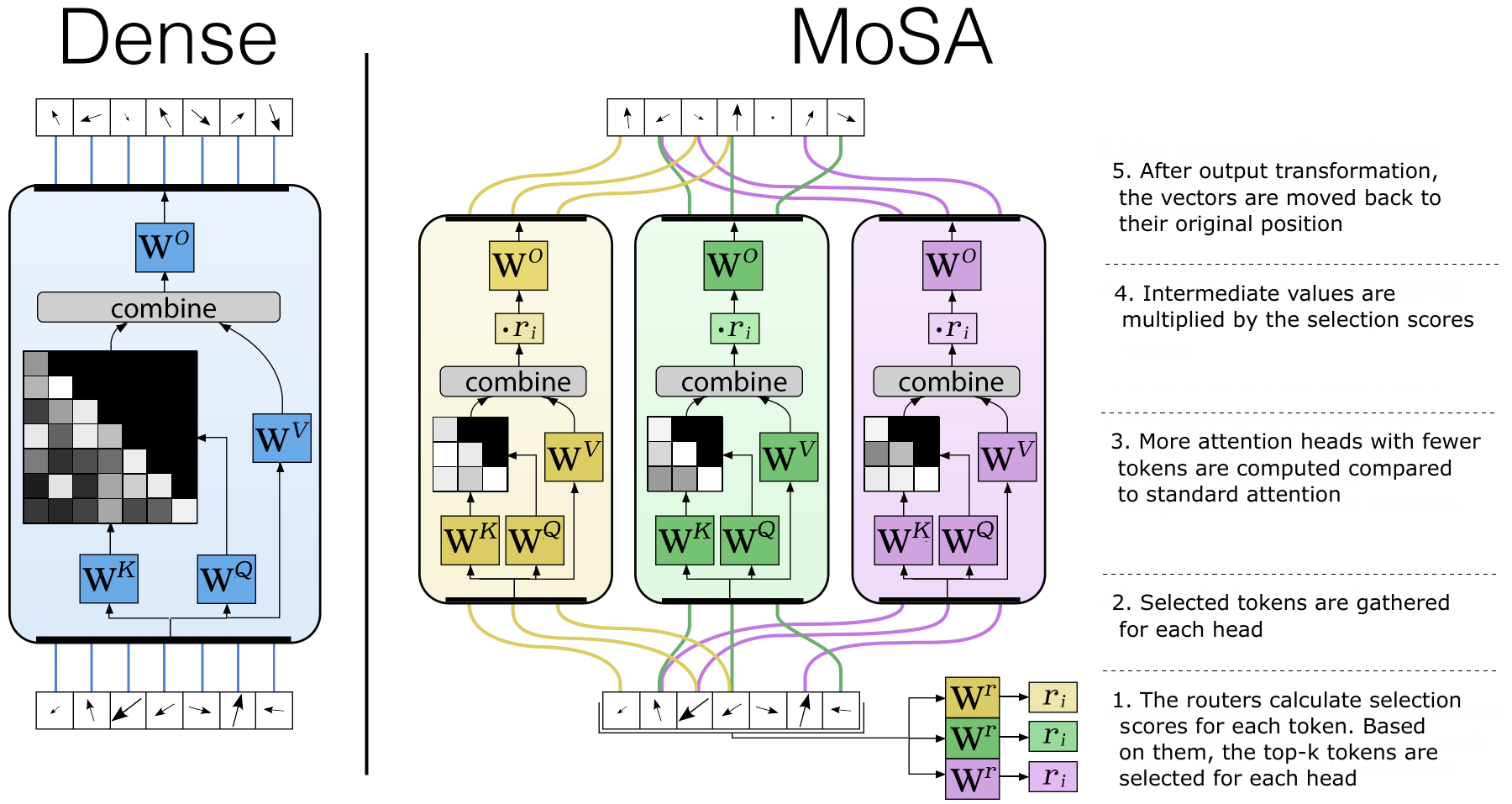}
    \caption{MoSA layer compared to the dense attention layer. MoSA replaces each dense head with multiple heads with a learnable sparsity pattern. Each head selects its own $k$ tokens to process. MoSA calculates query, key, and value projections only for the selected token and computes the attention only between them. It drops the rest of the tokens, leading to more efficient compute utilization. This reduces the computational and memory complexity on a sequence of length $T$ from $O(T^2)$ to $O(k^2+T)$. The saved compute budget can be used to scale up the number of heads.} 
    \label{fig:MoSA}
\end{figure}

As an alternative, static sparse attention methods \cite{child2019generating} reduce the quadratic complexity by selectively attending to a subset of tokens to be used in the attention. They use hand-defined coarse-grained patterns that are not data-dependent. Typical examples of these methods are the block-sparse and strided attention~\cite{child2019generating,zaheer2020big, beltagy2020longformer}. 
Static sparsity and block aggregation methods, however, impose significant limitations. They encourage the compression of multiple tokens into a single, lossy representation. This is necessary to remember information beyond the active block.
Such compression makes fine-grained recall difficult. The problem is similar to the well-known limitation of state-space models, which are forced to compress the entire past into a fixed-size representation
~\cite{arora2023zoology, jelassi2024repeat}. Content-based dynamic sparse attention~\cite{tay2021synthesizer, vyas2020fast, roy2021efficient} methods can, in principle, learn to attend to individual tokens, regardless of their location in the input, while ignoring less useful tokens. The Routing Transformer~\cite{roy2021efficient} clusters the tokens within each head using online K-means. However, it fails to show significant performance gains over static sparse-attention methods, possibly due to the slow convergence of online K-means ~\cite{bottou1994convergence}.

We propose a novel approach, inspired by Mixture-of-Experts \cite{shazeer2017outrageously, fedus2022switch}, to create a dynamic, content-based, and head-specific selection of tokens for sparse attention. This is achieved with Expert-Choice Routing~\cite{zhou2022mixture}, where each attention head is treated like an expert and selects its own specific tokens from the input. This creates a perfectly balanced selection, avoiding the need for complicated regularization techniques. We name our approach Mixture-of-Sparse Attention(\emph{MoSA}).
Although recent work explored applying ideas from the MoE literature to attention mechanisms \cite{zhang2022moa, csordas2023switchhead}, they focus on reducing the number of materialized dense attention matrices. We propose a different approach: we make the attention matrices sparse by selecting a small subset of tokens to use for each attention head. 

By selecting~$k$ tokens from a sequence of length~$T$, MoSA reduces the computational complexity of the attention head from $O(T^2)$ to $O(k^2+T)$. 
Sparse attention techniques have historically been employed out of necessity to manage long sequences that exceed available computational capacities. In contrast, we also explore the use of the saved computation budget for creating additional attention heads.
Thus, in this setup, MoSA employs a large number of highly sparse attention heads, encouraging their specialization.
We show that this allows for better utilization of the available compute budget and leads to substantially better iso-flop language modeling performance compared to dense attention. Furthermore, we analyze other sparse attention methods, such as fixed sparse attention~\cite{child2019generating} and the Routing Attention (the attention introduced in the Routing Transformer)~\cite{roy2021efficient}. MoSA is the only sparse attention method we analyzed that demonstrates improvement over dense baselines in the IsoFLOP setting.

Our main results demonstrate that hybrid models with many MoSA and four dense heads significantly improve the model's quality by up to $27\%$ in an IsoFLOP setting. Specifically, we evaluate MoSA on a language modeling task by starting with dense baselines and incrementally sparsifying the attention. We ensure FLOP-matching by swapping a specific number of dense heads for \emph{more} sparse heads. We repeat this procedure on different scales, starting with baselines from 28M to 516M parameters. MoSA consistently improves perplexity across all model scales. 

The IsoFLOP results demonstrate
MoSA's superior performance in a FLOP-matched setting.
However, sparse attention methods are often used to reduce computational and memory requirements. Furthermore, the idealized FLOP requirements often do not reflect wall-clock time. To demonstrate MoSA's efficiency, we show that in a perplexity-matched setting, MoSA exhibits both improved wall-clock time and GPU memory consumption even without a specialized CUDA kernel. It also reduces the total number of keys and values used in the computation, resulting in a significantly smaller KV cache. KV-cache size is an important practical problem for LLM inference and is the main focus of many post-training sparse attention methods \cite{liu2023scissorhands, li2025snapkv, zefan2024pyramidkv}. We investigate practical benefits in Section \ref{sec:resources}. 

In summary, our contributions are the following: \begin{enumerate}
    \item 
    We propose MoSA, a sparse attention method that uses a 
    \emph{learned}, context-based token selection, with each of the heads attending to a small subset of all tokens.  
    \item We evaluate MoSA in an IsoFLOP setting on four different scales with dense baselines ranging from 28M parameters to 516M. In this setting, MoSA improves perplexity by up to $27\%$. MoSA is the only sparse attention method we analyzed that improved perplexity compared to the dense baseline.
    \item We demonstrate that, in a perplexity-matched setting, a pure PyTorch implementation of MoSA improves both wall-clock time and memory usage simultaneously, without requiring specialized fast kernels. This setup also drastically reduces the KV cache size by using only a small subset of keys and values.
    \item We demonstrate that on long sequences, MoSA maintains a large advantage compared to other tested sparse-attention methods.
\end{enumerate}

The paper is organized as follows. Section~\ref{sec:method} provides the necessary background and describes MoSA. Section~\ref{sec:experiments} includes the experimental setup and the results. Section~\ref{sec:related_work} discusses related work, and Section~\ref{sec:future_work} provides a general discussion as well as potential future directions.

\section{Method}\label{sec:method}

\subsection{Background}

Here, we will discuss the necessary background on multi-head self-attention and mixtures of experts, which are required to understand our method.

\paragraph{Attention Mechanism.}

Attention assigns input-dependent weights to tokens in a sequence, allowing each token to gather context from the rest of the sequence. To do this, each token is projected to three vectors: its \textit{query}, \textit{key}, and \textit{value}. For a given token, we compare its \textit{query} vector with the \textit{key} vectors of all tokens (including itself), producing a set of similarity scores. The scores are then normalized and used to calculate a weighted sum of the tokens’ \textit{value} vectors. The result is a new representation that dynamically integrates information throughout the sequence.

Let $T$ be the sequence length, $h$ the hidden dimension of the model, and $h'$ the hidden dimension in each head. $\m{Q},\m{K},\m{V} \in \mathbb{R}^{T \times h'}$ represents the query, key and value matrices, respectively.

The attention output is computed as:
\begin{equation}
   \text{Attention}(\m{Q},\m{K},\m{V},\m{M}) = \text{softmax}\left(\frac{\m{QK}^\top + \m{M}}{\sqrt{h'}}\right)\m{V}\label{eq:attention}
\end{equation}
Here, $\m{M}$ denotes the attention mask that represents hard modeling constraints. $\m{M}_{i,j}=0$ if and only if $i'th$ token is allowed to attend to $j'th$ token, otherwise $\m{M}_{i,j}=-\infty$. In causal language models, $\m{M}_{i,j} = 0\iff i \geq j$ ensures that no token can attend to the future.

The multi-head attention (MHA) creates multiple instances of query, key, and value matrices from an input sequence $\m{X} \in \mathbb{R}^{T \times h}$ and applies the attention to each instance independently. These instances are called heads. Each head has its own mappings $\m{W}^Q_i, \m{W}^K_i, \m{W}^V_i \in \mathbb{R}^{h \times h'}$ and $\m{W}^O_i\in\mathbb{R}^{h' \times h}$, where $i \in \{1..H\}$ and $H$ is the number of heads. $h'$ is typically set to $\frac{h}{H}$. $\m{Q}_i=\m{X}\m{W}^Q_i, \m{K}_i=\m{X}\m{W}^K_i, \m{V}_i = \m{X}\m{W}^V_i$.
\begin{equation}
    \m{X}_{out} = \sum_{i=1}^H \text{Attention}(\m{Q}_i,\m{K}_i,\m{V}_i,\m{M}) \m{W}^O_i \label{eq:mha}
\end{equation}
The resulting mechanism allows the model to adaptively focus on relevant information while maintaining differentiability.
The lack of recurrence in the operations enables parallel processing of sequence elements. However, $\mathbf{QK}^\top$ is a $T \times T$ matrix and therefore introduces quadratic computational and memory complexity as a function of the sequence length.

\paragraph{Mixture of Experts.}
Mixture of Experts (MoE) combines multiple specialized neural networks (experts) with a gating mechanism that learns to route each input to the best-matching experts, activating only a small subset of experts per example. An MoE layer then computes its output as a sparsely weighted combination of the predictions of selected experts, with routing weights dynamically determined by the gating network.

Formally, given an input $\vx \in \mathbb{R}^h$, the MoE layer with $E$ experts and a scoring function (a router) $sel: \mathbb{R}^h\rightarrow\mathbb{R}^n$ can be expressed as $y(\vx) = \sum_{i\in \mathcal{E}} r_i(\vx)E_i(\vx)$
where $y(\vx)$ is the final output of the layer and $E_i(\vx)$ is the output of the expert $i$. $\mathcal{E}$ is the set of selected experts, usually defined as $\mathcal{E} = \text{argtopk}(r(\vx) + \varepsilon, k)$, where $k \in \mathbb{N}$ is the number of active experts, $\varepsilon$ is a stochastic noise present only during the training for exploration. The inputs are processed only by the active experts.

A critical challenge in MoE routing is ensuring balanced expert utilization. Without explicit constraints, the routing mechanism tends to overutilize a small subset of experts while leaving others largely inactive. This phenomenon, known as the load-balancing problem \cite{shazeer2017outrageously}, can significantly limit the capacity of the model and the effective number of parameters. Traditional approaches address this through auxiliary load-balancing losses \cite{lepikhin2020gshard, fedus2022switch} that encourage uniform expert utilization across a batch of inputs.

In contrast, Expert-Choice routing~\cite{zhou2022mixture} ensures perfect load balancing by inverting the traditional routing paradigm. Instead of the tokens choosing their experts, the experts choose which inputs they prefer to process. Given a batch of $B$ inputs, each expert selects the top-$k$ out of the $B$ inputs it will process.\footnote{In our case, each expert selects top-$k$ tokens from the sentence to process independently for each batch.}

Similarly to token choice routing, expert choice also reduces the \emph{average} number of experts used to process a token.
However, in contrast to token-choice routing, the amount of compute assigned is \emph{different} between tokens. This can be beneficial, as some tokens might be harder than others and therefore should benefit from more compute. On the other hand, it might lead to uneven resource allocation, where some tokens are assigned disproportionately high compute while others might starve.

Traditionally, Mixture of Experts has been applied in transformers as a replacement for a feedforward block, which is the most parameter-heavy part of the model. However, MoEs are sometimes also applied to the attention layer. SwitchHead~\cite{csordas2023switchhead} reduces the total number of heads by replacing some of the transformations with MoEs inside the attention. MoA~\cite{zhang2022moa} enhances Multi-Query Attention~\cite{shazeer2019fast} by adaptively selecting query transformations for each token. Similarly to these, we also apply the MoE to the attention layer, but in a way that introduces sparsity in the attention mechanism. 

\begin{figure}
    \centering
    \includegraphics[width=1\linewidth]{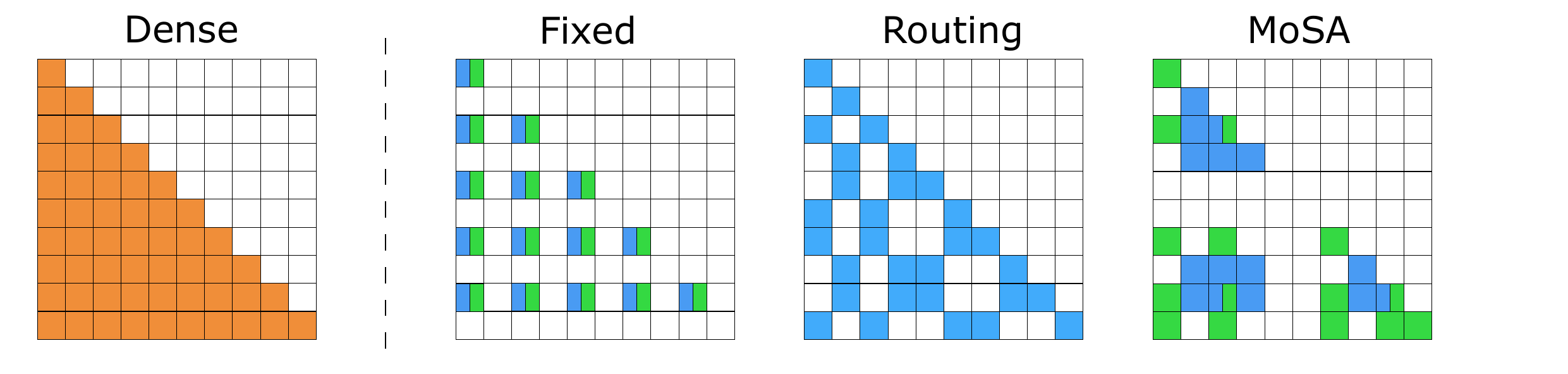}
    \caption{Attention variants visualized. In the plot, the colors indicate different heads. Sparse attention methods are roughly FLOP-matched and have sparsity $\rho=2$. One Routing Attention head corresponds in FLOP-cost to $\rho$ Fixed/MoSA heads. Fixed sparse attention uses only $k=\frac{T}{\rho}$ tokens in specific positions, with regular stride. The Routing Attention clusters tokens within each head into $\rho$ clusters of size $k$ based on their representations. MoSA selects $k$ tokens for each attention head independently based on their representations.  
    }
    \label{fig:attention-types}
\end{figure}

\subsection{Mixture of Sparse Attention (MoSA)}\label{sec:mosa}

Sparse attention methods model global dependencies by selecting specific tokens that can attend to other specific tokens based on a hand-engineered set of rules ~\cite{beltagy2020longformer, zaheer2020big} or by blockwise aggregation of tokens~\cite{yuan2025native}. Both of these families of methods impose the mixing of information during token aggregation, either explicitly or implicitly.

We propose instead to select tokens adaptively for each head based on the input. 
Thus, a flexible set of important tokens can be kept around, creating content-based sparsity without the need for information mixing. To achieve that, we take inspiration from Expert-Choice routing in MoEs.
We name our method \emph{Mixture of Sparse Attention (MoSA)}. MoSA learns which individual tokens to use for attention through end-to-end training. Each attention head in MoSA learns its own unique sparsity pattern, allowing different heads to specialize in different subsets of tokens relevant to their particular function within the network. This diverse, head-specific token selection pattern ensures that the model preserves the granular information within each relevant token while dynamically discovering optimal sparsity patterns specific to the data distribution. 
The architectural difference between MoSA and dense attention is illustrated in Fig. \ref{fig:MoSA}.
 
 The sparsity in MoSA reduces the computational cost of each attention head, allowing the use of more heads to develop targeted projections optimized for specific relationship types. 
The computational savings are particularly substantial when the number of selected tokens is significantly smaller than the sequence length.

In MoSA, in addition to the standard projections, each head has an additional router that selects which tokens are used for that head. Formally, the router is defined using the weight matrix $\m{W}^r \in \mathbb{R}^{h}$.
Let $\m{X} \in \mathbb{R}^{T \times h}$ be the $T$-long sequence of input tokens. The router calculates the selection scores for each token $\m{r} = \sigma(\m{X} \m{W}^r) \in \mathbb{R}^T$. For $\sigma$ we use the non-competitive sigmoid function $\sigma(x) = \frac{1}{1 + e^{-x}}$ following observations from $\sigma$-MoE \cite{csordas2023approximating}. Subsequently, we use expert choice for the selection of tokens for each head:
\[ \m{r}^{topk}, I = TopK(\m{r}, k)\]
where $TopK$ returns the highest $k$ values of $\vr$ called $\m{r}^{topk} \in \mathbb{R}^k$, along with their indices $I\in \{0,...,T-1\}^k$. $I$ is used to select the subset of inputs for the MoSA head:
\[\m{X}^s = (\m{X}_{I_1}, \m{X}_{I_2}, ... , \m{X}_{I_k}) \in \mathbb{R}^{k \times h}\]
where $\m{X}_i$ represents $i'th$ row from matrix $\m{X}$. After that, queries, keys, and values are calculated identically to the standard MHA: $\m{X}^s$ as $\m{Q}=\m{X}^s\m{W}^Q, \m{K}=\m{X}^s\m{W}^K, \m{V}=\m{X}^s\m{W}^V$. As our primary target is language modeling, we also calculate the mask that prohibits attending to future tokens. Unlike the standard MHA, this mask is not triangular and has to take into account the token indices selected by the head: $\m{M}_{i,j} = 0 \iff I_i \geq I_j, -\infty \text{ otherwise}$. 

The sparse attention can be computed using the standard attention defined in Eq.~\ref{eq:attention}. $\m{A}=\text{Attention}(\m{Q},\m{K},\m{V},\m{M})$. This allows the combination of MoSA with optimized attention implementations such as Flash Attention~\cite{dao2022flash}. The resulting vectors $\m{A}_i$ are multiplied by the corresponding router values $\m{r}_i$. Then, after the output transformation $\m{W}^o$, they are moved back to their original positions in the full-length sequence $\m{Y} \in \mathbb{R}^{T \times h}$. 
\begin{align*}
\m{X}^o &= \operatorname{diag}(\m{r})\m{AW}^o  \in \mathbb{R}^{k \times h}\\
\m{Y}_j &= \begin{cases}
\m{X}^o_i, & \text{if } j = I_i \text{ for some } i \in \{1, \dots, k\}, \\
0,     & \text{otherwise},
\end{cases} \quad \text{for } j = 1, \dots, T.
\end{align*}
 $\text{diag}(\cdot)$ creates a diagonal matrix from a vector, used for elementwise scaling of the columns of the matrix $\m{A}$ by a vector $\m{r}$. This ensures that the token's contribution is proportional to the router's output. This also enables the router to receive gradients, making it learnable by gradient descent.

We call the combined transformation of $x$ into $y$, parameterized by $\theta_i = (\m{W}^Q, \m{W}^K, \m{W}^V, \m{W}^O, \m{W}^r)$ a single MoSA head: $\m{Y} = \text{MoSA}_{head}(\m{X}; \theta_i)$. A MoSA layer parameterized by $\theta = \{\theta_i\}_{i\in{1...H}}$ is a sum of all MoSA heads
\begin{equation}
\text{MoSA}(\m{X}; \theta) = \sum_{i=1}^{H} \text{MoSA}_{head}(\m{X}; \theta_i) 
\label{eq:mosa_mha}
\end{equation}
The entire transformation in the multihead version can be efficiently implemented in PyTorch~\cite{paszke2019pytorch} using \texttt{einsum}, \texttt{scatter} and \texttt{gather} operations.

\paragraph{Hybridization.} Sparse attention methods are usually combined with local attention~\cite{child2019generating, roy2021efficient} when used on long sequences. Sparse attention then captures global dependencies, while local attention preserves local context. As our setup permits the use of dense attention, in our main experiments, we combine MoSA or corresponding sparse attention baseline with 4 dense heads. In Appendix~\ref{app:hybrid}, we demonstrate the necessity of hybridization and motivate our selection of four dense heads for the models. In Section~\ref{sec:long_sequences}, we combine MoSA with local attention for long sequences and demonstrate that MoSA demonstrates superior performance in this scenario as well.

\paragraph{Positional encodings.} All our experiments use Rotary Positional Encodings (RoPE)~\cite{su2021roformer}. RoPE applies positional encodings for each attention head after query and key mapping. It does this by rotating them at an angle determined by the token's position in a sentence. Similarly to the attention mask, we must ensure that the rotations correspond to the token's original position in the sequence $\m{X}$ rather than the selected subset $\mX^S$. Thus, we adapt RoPE to be aware of token positions $I$.
Following standard practice, we rotate half of the dimensions and leave the other half unchanged.

\section{Experiments}\label{sec:experiments}

\begin{table}[h]
    \centering
    \resizebox{\textwidth}{!}{
    \begin{tabular}{c|ccccc}
        \toprule
        Model size & \#Params Dense & Dense ppl $\downarrow$ & MoSA Best ppl $\downarrow$ & Fixed Best ppl $\downarrow$ & Routing Best ppl $\downarrow$ \\ 
        \midrule
        Tiny   & 28M  & 22.46 & 16.39 \textcolor{ForestGreen}{($-27.0\%$)} & 23.28 \textcolor{Maroon}{($+3.7\%$)} & 23.33 \textcolor{Maroon}{($+3.9\%$)} \\
        Small  & 113M & 16.01 & 12.85 \textcolor{ForestGreen}{($-19.7\%$)} & 16.51 \textcolor{Maroon}{($+3.1\%$)} & 16.43 \textcolor{Maroon}{($+2.6\%$)} \\
        Medium & 210M & 13.95 & 11.06 \textcolor{ForestGreen}{($-20.7\%$)} & 14.35 \textcolor{Maroon}{($+2.9\%$)} & 14.21 \textcolor{Maroon}{($+1.9\%$)} \\
        Large  & 516M & 12.20 & 10.58 \textcolor{ForestGreen}{($-13.3\%$)} & 12.40 \textcolor{Maroon}{($+1.6\%$)} & 12.24 \textcolor{Maroon}{($+0.3\%$)} \\ 
        \bottomrule
    \end{tabular}
    }
    \caption{Comparing
    dense and sparse models (Fixed, Routing, MoSA) under a fixed computational budget (see Section~\ref{sec:isoflop}). For sparse models, the table contains the best perplexity across all sparsities bigger than 1. The results for sparse models were selected as the best of all sparsities. Relative difference to the dense baseline is displayed in the parentheses. MoSA significantly outperforms the dense baseline, reducing perplexity by up to 27\%. The fixed and the Routing Transformer baselines both fail to reach the performance of the dense model.}
    \label{tab:main_results_comparison}
\end{table}

In this section, we empirically demonstrate MoSA's performance in different settings. We compare MoSA to dense and sparse baselines introduced in Section~\ref{sec:baselines}. In Section~\ref{sec:isoflop}, we evaluate all the methods on language modeling under a fixed FLOP budget. In Section~\ref{sec:resources} we demonstrate the practical benefits of MoSA by measuring wall-clock time, memory usage, and KV cache size in a perplexity-matched setup. In Section~\ref{sec:long_sequences} we investigate the performance of MoSA on long sequences. Finally, in Section~\ref{tab:downstream_tasks} we show the performance of different models in downstream zero-shot tasks.

We use four model sizes for our experiments:  \textit{Tiny}, \textit{Small}, \textit{Medium} and \textit{Large}. Each size is defined by the \emph{FLOP count} of the forward pass of the corresponding dense transformer baseline. The parameter count of dense models associated with each size is: 28M for \textit{Tiny}, 113M for \textit{Small}, 210M for \textit{Medium}, and 516M for \textit{Large}.

\paragraph{Implementation details}
We use the SentencePiece~\cite{KudoR18} tokenizer based on sub-word units~\cite{sennrich16bpe, SchusterN12} a vocabulary size of 8000.  
All our models are trained on the C4~\cite{raffel2020exploring} dataset for 100k batches, with batch size $B=64$ and sequence length $T=1024$. This means that we train on the $10^5SB \approx 6.5B$ tokens from the dataset.
We use the Adam~\cite{kingma2014adam} optimizer with a learning rate of 0.00025, gradient clipping above the norm of 0.25, and a linear warmup for 4k steps. For detailed hyperparameters, please refer to Appendix \ref{app:models_details}.

\subsection{Baselines}\label{sec:baselines}

Apart from a dense baseline, we compare MoSA with two sparse attention methods: static, position-based sparse attention, and content-based sparse attention. 

\paragraph{Fixed Sparse Attention.}
Position-based static attention patterns have been shown to be a strong
sparse attention
variant~\cite{child2019generating}, outperforming strided sliding window attention. Fixed sparse attention for a sparsity $\rho$ selects $k=\frac{T}{\rho}$ tokens with stride $\rho$. Using the notation introduced in Section~\ref{sec:mosa}, fixed sparse attention can be written as a special case of MoSA, where $I = [0,\rho,2\rho,...,T-\rho]$ and $\m{r}=\m{1}$.

Fixed sparse attention reduces computational complexity in two ways. First, it decreases the $O(T^2)$ cost of the full attention matrix by limiting attention to predefined token positions. Second, since only these pre-selected tokens participate in attention calculations, the query, key, value, and output transformations need only be computed for this subset rather than all tokens.

However, this approach introduces information flow constraints. Pre-selected tokens must aggregate necessary information in earlier layers. Furthermore, in the subsequent layers they have to be routed back to the positions where they are most useful. This additional overhead in information routing limits the model's representational capacity and overall expressiveness.

\paragraph{The Routing Transformer.}
We also compare MoSA to the content-based attention proposed in the Routing Transformer~\cite{roy2021efficient}. The Routing Attention is the most similar method to MoSA we found in the literature. It groups tokens with online K-means into $\rho$ clusters of size $k=\frac{T}{\rho}$  \emph{inside} each head. This is implemented during training by the top-k tokens most similar to the cluster centers using the dot-product distance metric. 
Cluster centers are learned using a moving average of the most similar tokens.

The Routing Attention might resemble the Expert-Choice selection with MoSA. There are, however, several crucial differences that, as our experiments show, lead to significant differences in the performance of MoSA in comparison to the Routing Transformer. Specifically, online K-means, used for clustering in the Routing Transformer is known for suffering from an extremely slow convergence rate~\cite{bottou1994convergence}. 
It is also unclear if clustering keys and queries is well aligned with the language modeling objective.
In contrast, the learned dynamic matching mechanism of MoSA is directly optimized by the same objective as the model.

MoSA benefits from the sparsity in the $\m{W}^Q, \m{W}^K,\m{W}^V,\m{W}^O$ transformations, which need to be computed only for selected tokens.
 In contrast, the Routing Transformer has to compute all keys and queries before the clustering step. 
MoSA's efficiency enables the use of more heads with specialized weights in a smaller subset of tokens. 
Its selection can also lead to dynamic compute allocation, where some more important tokens are processed by more heads than less important tokens.

Last but not least, the Routing Transformer performs best in language modeling when the clusters share the same destination (query) tokens and source (keys and values) tokens. In our experiments, we also found that MoSA performs better if the same tokens are selected for the source and destination sides. However, to enforce this in the Routing Transformer, they require to set $\m{W}^Q=\m{W}^K$.
In MoSA, however, the same selection for source and destination side can be enforced with $\m{W}^Q$ different from $\m{W}^K$, allowing greater flexibility.

We visualize typical schematic attention patterns of the baselines and MoSA in Fig.~\ref{fig:attention-types}. Note that several previous works proposed combining different types of sparse attention to achieve synergic performance in long-sequence tasks~\cite{beltagy2020longformer, zaheer2020big, zhang2023efficient}. In this work, we focus on investigating sparse attention methods in combination with a few dense attention heads, but without combining multiple sparse attention types. We leave combining MoSA with other sparse-attention methods for future work.

\subsection{Main Results}\label{sec:isoflop}

\begin{figure}
    \centering
    \includegraphics[width=1\linewidth]{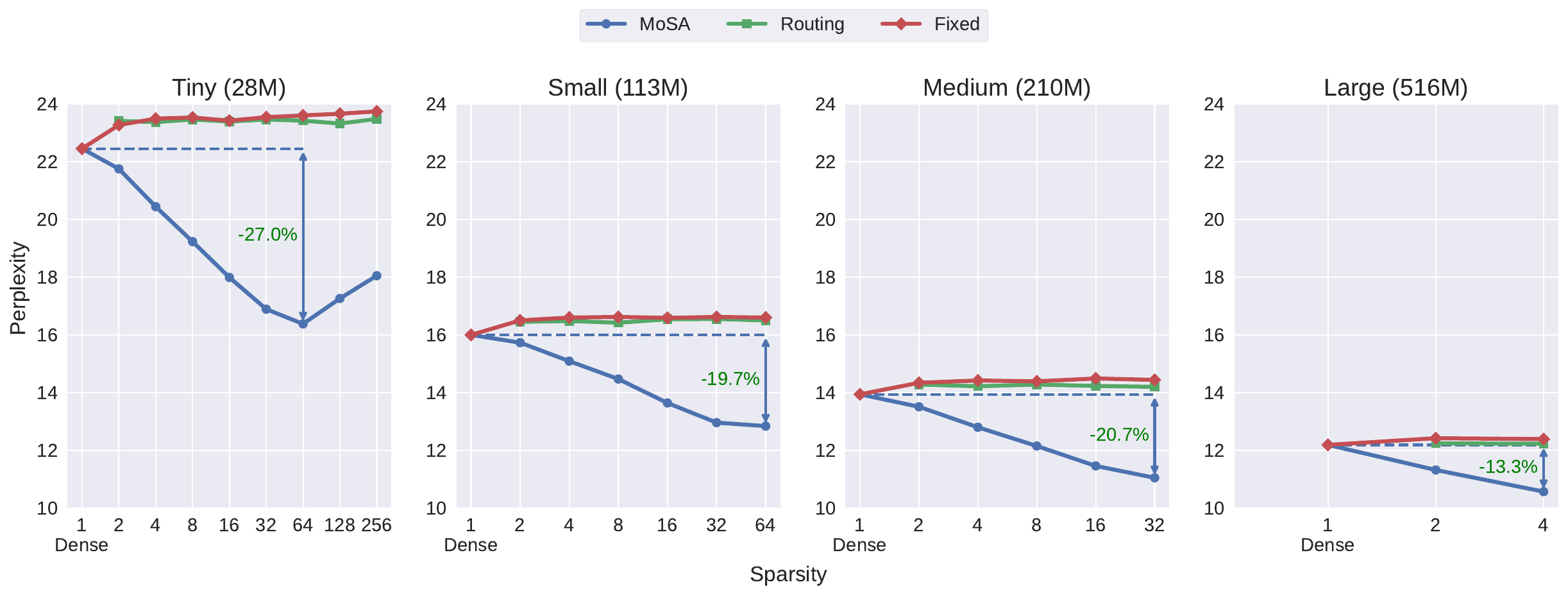}
    \caption{Perplexity ($\downarrow$) of FLOP matched models under different sparsities. Each plot corresponds to a specified FLOP budget per step. The number in parenthesis is the number of parameters of the dense baseline. Sparsity 1 represents the dense baseline. As sparsity increases, MoSA's perplexity improves monotonically until reaching a saturation point around sparsity 32-64, beyond which performance deteriorates.
    This is likely because at very high sparsity levels, each attention head selects only a few tokens, which is insufficient to capture the complex relations. 
    On the other hand, other sparse methods fail to reach the perplexity of the dense baseline in the IsoFLOP setting. We explore fewer sparsity levels for larger models due to excessive memory requirements.}
    \label{fig:main_results}
\end{figure}

To evaluate sparse methods, we evaluate multiple models with a gradually increasing sparsity rate $\rho=\frac{T}{k}$.
This reduces the compute requirements for each sparse head. We use the saved budget to increase the number of sparse heads. Specifically, we choose the number of sparse heads to be the maximum such that the FLOPs of the sparse model do not exceed the FLOPs of the baseline model for a given size. All sparse models include four dense heads that we keep (see Section~\ref{sec:mosa}), and are included in the FLOP calculations. 

Note that increasing the number of attention heads also increases the memory requirements of all methods. Consequently, for the larger FLOP-matched models, we restricted the explored sparsity values to ensure that the models fit in the memory budget dictated by our hardware.

In the IsoFLOP experiments, following observations from StreamingLLM~\cite{xiao2024efficient} on the importance of first tokens in the attention mechanism, we always include the first token in all MoSA heads. The head selects $k-1$ tokens based on their router scores and the first token. The representation of the first token, just like the others, is multiplied by its router score after the attention mechanism.

\paragraph{FLOPs Calculation.}\label{sec:flop_calculation}
Let $T$ be the sequence length, $h$ the hidden dimension of the model, $h'$ the hidden dimension in each head (after passing through the query, key or value projection), $k$ the number of tokens selected for each head, and the sparsity rate $\rho=\frac{T}{k}$.

FLOPs cost of a single head is equal to:
\begin{align*}
\text{FLOP}_{\text{dense}} &= \underbrace{\textcolor{blue}{8hh'T}}_{\text{Q,K,V,O mappings}} + \underbrace{\textcolor{orange}{4h'T^2}}_{\text{Attention}} \\[10pt]
\text{FLOP}_{\text{mosa}} &= \underbrace{\textcolor{blue}{8hh'k}}_{\text{Q,K,V,O mappings}} + \underbrace{\textcolor{orange}{4h'k^2}}_{\text{Attention}} + \underbrace{\textcolor{purple}{2hT + h'k}}_{\text{routing overhead}} \\ 
\text{FLOP}_{\text{fixed}} &= \underbrace{\textcolor{blue}{8hh'k}}_{\text{Q,K,V,O mappings}} + \underbrace{\textcolor{orange}{4h'k^2}}_{\text{Attention}} \\
\text{FLOP}_{\text{routing}} &= \underbrace{\textcolor{blue}{6hh'T}}_{\text{Q=K,V,O mappings}}+ \underbrace{\textcolor{orange}{4h'k^2\rho}}_{\text{Attention}} + \underbrace{\textcolor{purple}{2h'T}}_{\text{cluster selection}} = \rho(\textcolor{blue}{6hh'k} + \textcolor{orange}{4h'k^2} ) + \textcolor{purple}{2h'T}  \\
\end{align*}
The detailed derivation of FLOP costs of the attention and the entire models can be found in App.~\ref{app:FLOPs}. Note that, typically $k << T$, hence the MoSA head is significantly cheaper compared to a dense head.

The selection mechanism in MoSA introduces an additional overhead of $2hT + h'k$ ($2hT$ comes from token scoring and $h'k$ comes from multiplying the output by the scores), which is small compared to the rest. As a consequence, the cost of the MoSA head is comparable to that of the fixed sparsity attention head, while allowing content-based dynamic sparsity.

In contrast to MoSA and fixed attention, the Routing Transformer must compute all tokens by query, key, value, and output transformations. However, in the Routing Transformer for autoregressive text $K=Q$, therefore, only 3 projections need to be computed. Hence, the projection cost is equal to $6hh'T$. The attention in the Routing Transformer has multiple clusters inside each head. More specifically, it has $\rho$ clusters of size $k$, and therefore the attention cost of the head is equal to the attention cost of the cluster multiplied by the number of clusters. The Routing Transformer has an additional layer normalization inside the head, which we omitted for simplicity.

FLOP-wise, one Routing Attention head more or less corresponds to $\rho$ fixed attention or $\rho$ MoSA heads. Loosely speaking, MoSA with $\rho$ heads is similar to the Routing Attention head, where each cluster has its own custom linear transformation, rather than a single one shared among clusters.

\paragraph{IsoFLOP Curves.}

Starting from sparsity 1, which corresponds to the dense model, we gradually increase the sparsity and measure the test-set perplexity of FLOP-matched models. Table~\ref{tab:main_results_comparison} lists the best results for each model class and size. Across all model sizes tested, MoSA achieved significantly better perplexity within fixed FLOP budgets compared to dense baselines. All MoSA hybrids reduce the perplexity of the baseline, sometimes by $27\%$. On the other hand, the sparse baselines for all sparsities $\rho > 1$ perform worse 
than the dense baseline. 

Figure~\ref{fig:main_results} illustrates the IsoFLOP curves of the models with varying degrees of sparsity. For MoSA, performance steadily improves as sparsity increases, reaching optimal results at approximately $\rho=64$. Beyond this threshold, performance begins to decline, creating a "U" shape in the curve. 
This is likely because the excessively high sparsity values limit the model's ability to capture complex attention patterns. For example, at $\rho=256$ with a sequence length of $T=1024$, only $k=4$ tokens are selected to participate in each attention head.

For some configurations, MoSA turns proves to be more efficient than the dense model even in a parameter-matched setting. For example, \textit{Medium} model with sparsity $8$ has $442M$ parameters and perplexity $12.16$, while the \textit{Large} baseline model has $516M$ parameters and perplexity $12.20$. This shows that a higher specialization of the heads might lead to improved performance even when we discard computational benefits. Detailed results for different MoSA sparsity configurations, together with the total number of parameters and the number of heads, are listed in the Appendix~\ref{app:detailed_isoflop_results}.

In contrast to MoSA, both fixed sparse attention and the Routing Attention consistently underperform the dense baseline across all sparsity levels. They exhibit relatively constant, but worse, perplexity across different sparsity values, with only minor fluctuations that reveal no discernible trend.

\subsection{Resource Optimization} \label{sec:resources}

The previous section demonstrates MoSA's ability to achieve better perplexity than dense transformers with an identical compute budget. In this section, we examine MoSA's practical efficiency gains. Specifically, we match the perplexity scores between the MoSA and the dense baseline to measure wall-clock time, memory, and KV-cache size savings. 

To find the perplexity-matched comparison, we select sparsity to be equal to 32 for model sizes \textit{Tiny, Small} and \textit{Medium}. For \textit{Large} we select $\rho=16$ to keep sparsity closer to the range investigated in Section~\ref{sec:isoflop}. Then, we gradually increase the number of MoSA heads until the perplexity matches the dense baseline. We do it for all four model scales defined in Section~\ref{sec:isoflop}.

The results are shown in Table~\ref{tab:resource_optimization}.
MoSA can match the dense baseline, while being faster in wall-clock time and using less memory at the same time. These findings show that MoSA not only improves model quality in the FLOP-matched setting but can also be used to reduce computational and memory requirements when targeting the same performance level. Furthermore, it shows that MoSA uses computation more effectively than standard dense attention across all efficiency metrics. 

MoSA achieves this without a specialized CUDA kernel using only PyTorch-level operations. We expect that designing a specialized kernel would result in additional significant efficiency gains.

In addition to the speed and memory used for the training, we report the total number of key-value pairs (KV) used, calculated as $\text{KV} = TH_{dense} + kH_{mosa}$, where $H_{dense}$ and $H_{mosa}$ represent the number of dense and sparse heads, respectively. 
KV directly corresponds to the size of the costly KV-Cache in the autoregressive setting. KV cache optimization has been the goal of many post-training sparse-attention methods\cite{liu2023scissorhands, li2025snapkv, zefan2024pyramidkv}. Our results demonstrate that MoSA offers a significant reduction in KV-cache size while simultaneously improving speed and memory requirements.

\begin{table}[h!]
\centering
\resizebox{\textwidth}{!}{
  \begin{tabular}{c||cc|cc|cc|cc}
  
\toprule
   & \multicolumn{2}{c|}{Tiny} & \multicolumn{2}{c|}{Small} & \multicolumn{2}{c|}{Medium} & \multicolumn{2}{c}{Large} \\

     & Dense  & MoSA & Dense & MoSA & Dense & MoSA & Dense  & MoSA   \\
  \midrule
  Dense Heads            
         & 9  & 4  & 9  & 4  & 9  & 4  & 16 & 4  \\
  MoSA Heads             
         & 0  & 17 & 0  & 14 & 0  & 12 & 0  & 16 \\
  Perplexity ($\downarrow$)           
         & 22.46 & 22.40  & 16.02 & 16.01  & 13.94 & 13.76  & 12.20  & 12.16  \\
  Wall-time/step $\downarrow$ (ms) 
         & 137   & \textbf{127}    & 326   & \textbf{319}    & 619   & \textbf{592}   & 807    & \textbf{703}    \\
  Wall-time/step gain (\%) 
         & --    & \textcolor{ForestGreen}{$-7.3\%$}  & --    & \textcolor{ForestGreen}{$-2.1\%$}  & --    & \textcolor{ForestGreen}{$-4.4\%$}  & --    & \textcolor{ForestGreen}{$-12.9\%$} \\
  Memory $\downarrow$ (GB)      
         & 21.1  & \textbf{19.0}  & 32.4  & \textbf{31.4}  & 50.2  & \textbf{49.4}  & 104.1 & \textbf{94.5}  \\
  Memory gain (\%)      
         & --    & \textcolor{ForestGreen}{$-10.0\%$} & --    & \textcolor{ForestGreen}{$-3.1\%$}  & --    & \textcolor{ForestGreen}{$-1.6\%$}  & --    & \textcolor{ForestGreen}{$-9.2\%$}  \\
  KV Total $\downarrow$ (K)
         &    9.2   &   \textbf{4.5}   &   9.2    &   \textbf{4.4}    &    9.2   &     \textbf{4.4}  &    16.4   &    \textbf{5.0}   \\
 KV Total gain (\%)    
         & --    & \textcolor{ForestGreen}{$-51.1\%$} & --    & \textcolor{ForestGreen}{$-52.2\%$} & --    & \textcolor{ForestGreen}{$-52.2\%$} & --    & \textcolor{ForestGreen}{$-69.5\%$} \\
 
  \bottomrule
  \end{tabular}
}
\caption{Resource usage reduction from perplexity-matched MoSA models. KV is the KV-cache size, representing the total number of key-value pairs required (in thousands). MoSA models match the perplexity of dense baselines while at the same time improving wall-clock time, using less memory, and significantly smaller KV cache for all model sizes. Resource usage was measured on a single A100 GPU for \textit{Tiny, Small} and \textit{Medium} models and on two A100 GPUs for \textit{Large}.}
\label{tab:resource_optimization}
\end{table}

\subsection{Scaling with Sequence Length}\label{sec:long_sequences}

Traditionally, sparse attention methods have been introduced as a necessity when sequence length makes dense attention computationally prohibitive. 
After demonstrating MoSA's effectiveness in standard-length sequences, we now investigate whether MoSA's benefits are retained or amplified in this long sequence setup. 

In contrast to previous sections, here we combine MoSA or a baseline method with local attention~\cite{child2019generating, beltagy2020longformer}. We use local attention instead of dense attention because even a small number of dense attention heads would result in prohibitive memory usage in a longer context scenario. This is a standard practice in the sparse attention literature~\cite{child2019generating, roy2021efficient}. Local attention preserves local dependencies, while global, sparse attention enables efficient processing of long dependencies.

We scale our sequence length from 1024 to 8192 tokens and keep the $k$ constant equal to $64$. Hence, the sparsity increases from $\rho=16$ for $T=1024$ to $\rho=128$ for $T=8192$. Contemporary sparse attention methods for long sequences are trained in longer sequences~\cite{yuan2025native}. However, due to our limited hardware budget, we restrict our experiments to a sequence length of $8192$. We treat this investigation as a preliminary analysis that demonstrates the potential of MoSA for long sequences. Importantly, it demonstrates that MoSA performs well when combined with local attention, which is a typical long-sequence setup. 

As in the previous section, we compare MoSA with fixed sparse attention and the Routing Attention. All long sequence models have 6 layers and hidden dimension size of 1024. The Routing Transformer has 4 local attention heads and 4 Routing Transformer heads in all layers, whereas the fixed sparse attention and MoSA have 60 sparse heads and 4 local attention heads. We chose 60 sparse heads to roughly FLOP match all models for $T=1024$. However, as we keep $k$ constant, for longer sequences with $2048, 4096$ and $8192$ tokens, the FLOP cost for fixed attention and MoSA will be much lower than for the Routing Attention. For $T=8192$ FLOP cost of $60$ MoSA's heads is equal to only $22.99\%$ of $4$ Routing Transformer heads. 

\begin{figure}
    \centering
    \includegraphics[width=0.75\linewidth]{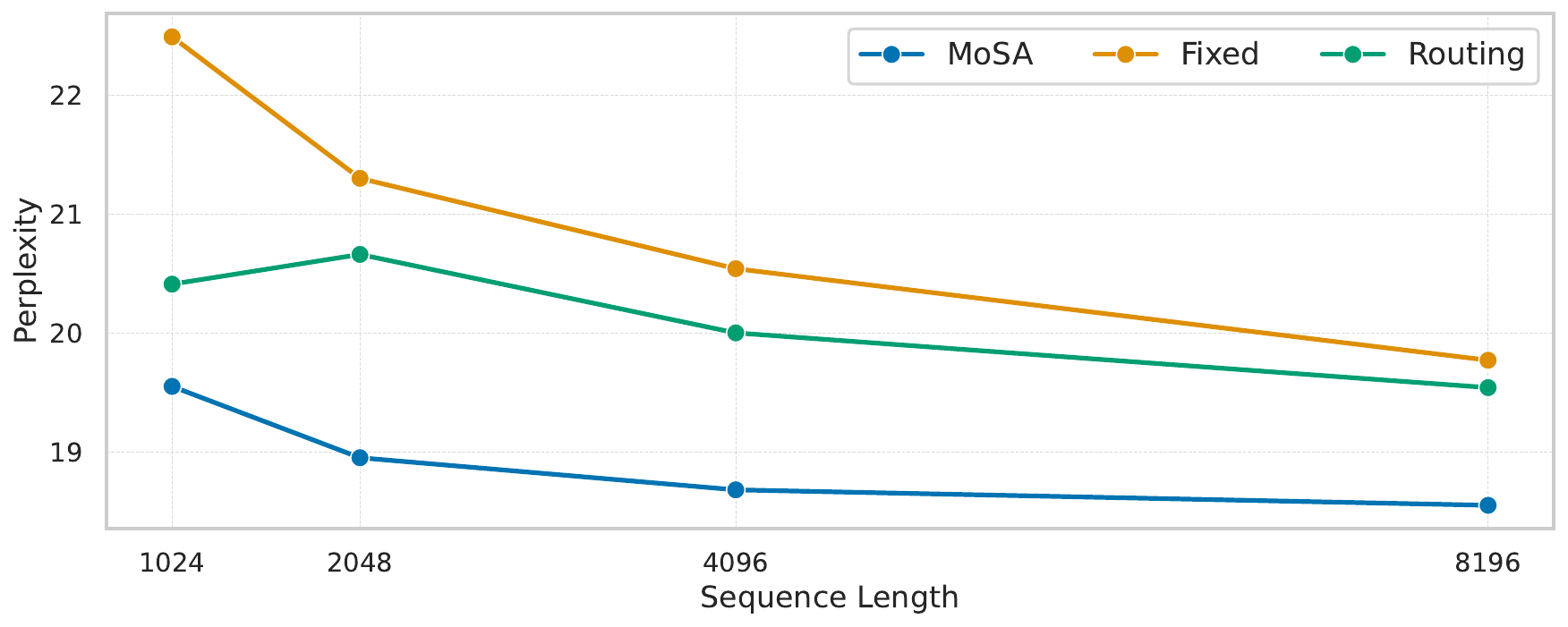}
    \caption{Perplexity of sparse‑attention methods (MoSA, Fixed, and Routing) as sequence length increases. Each method has a fixed size window size (cluster size for the Routing Transformer, number of tokens selected for each head in MoSA and Fixed) regardless of total sequence length. MoSA matches the computational cost of the fixed sparsity baseline while requiring fewer FLOPs than the Routing Attention and consistently achieves the lowest perplexity.}
    \label{fig:long_sequences}
\end{figure}

The results are shown in Fig.~\ref{fig:long_sequences}. MoSA significantly outperforms other sparse attention methods across all sequence lengths. This is true even at length $8192$, where MoSA uses only a small fraction of the computational cost of the Routing Transformer.

The significant performance gap in the results demonstrates the potential of MoSA for ultra-long sequences~\cite{kitaev2020reformer, yuan2025native, xu2025128k}. Given our limited resources, we leave the investigation of MoSA in this context for future work.

\subsection{Downstream Tasks}\label{sec:downstream}

We evaluate the zero‑shot downstream performance of MoSA on six established benchmarks: LAMBADA~\cite{paperno2016lambada}, WinoGrande~\cite{sakaguchi2020winogrande}, BLiMP~\cite{warstadt2020blimp}, HellaSwag~\cite{zellers2019hellaswag}, PIQA~\cite{bisk2020piqa} and AI2ARC~\cite{clark2018think}—covering tasks from cloze‑style completion to commonsense reasoning. 

During training, MoSA operates on sequences of more or less constant size $T=1024$. However, for downstream tasks, some inputs will be much shorter. For example, most datapoints in the BLiMP dataset do not exceed 10 tokens. In order to handle such situations, we adaptively choose the number of tokens for each input to be $k = \max(\lfloor\frac{T}{\rho}\rfloor,2)$ tokens for each head. This simulates the ratio of tokens selected for the attention head during the training. Moreover, it ensures that at least $2$ tokens are selected, which is the minimum necessary for the attention to model any cross-token dependencies.

For each scale and sparse model type, we select the model with sparsity $\rho > 1$ that produced the best perplexity in the IsoFLOP scenario (Sec.~\ref{sec:isoflop}). We also include the dense baseline for each size. Table~\ref{tab:downstream_tasks} reports the performance across the tasks. The best result for a given task across model types is bold.

For \textit{Tiny}, \textit{Small}, and \textit{Medium} scales, MoSA generally outperforms other models. BLiMP stands as a notable exception, where MoSA consistently underperforms. This weak performance on BLiMP can be attributed to the extremely short length of most examples in the dataset. With longer sequences seen during training, each MoSA head can selectively process only the tokens it handles well. However, in short sequences, the shortage of tokens forces MoSA heads to operate on tokens outside their training distribution. Furthermore, when $\lfloor\frac{T}{\rho}\rfloor = 1$, resulting in only 2 tokens being selected, there is a significant discrepancy between the percentage of selected tokens compared to training conditions. Models with a high sparsity factor of $64$ typically select only $1.56\%$ tokens in a sequence for each attention head. Yet for a sequence length of $T=10$, $2$ selected tokens represent $20\%$ of the sentence, creating a distribution mismatch.

Moreover, in \textit{Large} scale, the Dense baseline outperforms MoSA despite having much higher perplexity. We attribute the downstream performance gap of MoSA to two main factors. First, MoE architectures have been shown to suffer from expert overspecialization, which often leads to decreased performance in downstream tasks~\cite{fedus2022switch, zoph2022st}. Instruction tuning has been shown to mitigate this issue~\cite{shen2024mixture}. 

Furthermore, content‑based sparse attention methods tend to struggle on shorter sequence\footnote{See: 
\url{https://github.com/lucidrains/routing-transformer?tab=readme-ov-file\#issues}}. Our experiments confirm this pattern, as MoSA outperforms the Routing Attention in most tasks. Furthermore, some runs of the Routing Attention were unstable in context of downstream tasks (Medium scale of the Routing Attention). 
Practitioners report that extending training by additional epochs on truncated sequences can mitigate the issues of sparse attention methods on short sequences\footnotemark[\value{footnote}].

\begin{table}[htbp]
\centering
\begin{tabular}{llcccccc}
\toprule
 & Model & LAMBADA & WinoGrande & BLiMP & HellaSwag & PIQA & AI2ARC \\
\midrule
\multirow{4}{*}{\textbf{Tiny}} 
  & Dense   & 18.7 & 50.3 & 72.0 & 27.5 & \textbf{59.4} & 28.0 \\
  & MoSA    & \textbf{25.4} & \textbf{51.9} & 64.6 & \textbf{29.1} & \textbf{59.4} & \textbf{28.6} \\
  & Routing & 14.0   & 51.3   & 66.2   & 27.8   & 57.1   & 25.9   \\
  & Fixed   & 17.1   & 50.6   & \textbf{72.5}   & 27.7   & 58.6   & 28.1   \\
\midrule
\multirow{4}{*}{\textbf{Small}} 
  & Dense   & 25.8 & \textbf{52.1} & \textbf{76.2} & 30.9 & 62.4 & 30.1 \\
  & MoSA    & \textbf{30.7} & 48.5 & 62.8 & \textbf{31.8} & 60.4 & \textbf{30.2} \\
  & Routing & 19.2   & 50.7   & 70.2   & 28.0   & 57.6   & 27.3   \\
  & Fixed   & 24.6   & 51.6   & 75.3   & 30.1   &  \textbf{63.2}  & \textbf{30.2}   \\
\midrule
\multirow{4}{*}{\textbf{Medium}} 
  & Dense   & \textbf{31.4} & 51.2 & \textbf{77.8} & 33.8 & 64.5 & 31.5 \\
  & MoSA    & 27.6 & \textbf{52.2} & 75.1 & \textbf{33.9} & \textbf{65.1} & \textbf{31.6} \\
  & Routing &  10.2  & 51.5   & 65.9   & 30.3   & 57.8   & 27.8   \\
  & Fixed   & 29.4   & 51.4   & 77.3   & 33.0   & 64.6   & 31.5   \\
\midrule
\multirow{4}{*}{\textbf{Large}} 
  & Dense   & \textbf{36.2} & 52.5 & \textbf{80.4} & \textbf{38.7} & \textbf{67.1} & \textbf{33.8} \\
  & MoSA    & 32.3 & \textbf{52.8} & 77.2 & 36.6 & 65.0 & 32.2 \\
  & Routing & 27.5   & 51.1   & 76.5   & 36.2   & 64.1   & 32.5   \\
  & Fixed   & 32.3   & 51.7   & 79.6   & 35.9  & 66.0   & 32.2   \\
\bottomrule
\end{tabular}
\caption{Accuracy on downstream zero-shot tasks. Each model is selected with the best sparsity in the IsoFLOP comparison. Note that on downstream tasks, the token selection mechanism of MoSA operates out of distribution. Despite this, MoSA often outperforms the dense baseline. Even when it doesn't, the performance gap is usually small.}
\label{tab:downstream_tasks}
\end{table}

\section{Related Work}\label{sec:related_work}

The quadratic cost of attention in the 2017 transformer model \cite{vaswani2017attention} has led to a wide body of research on efficient attention variants \cite{kitaev2020reformer, choromanski2021rethinking}. Popular alternatives are different linear attention variants that typically use a fixed vector or matrix memory and update it recurrently. The 1992 unnormalised linear Transformers \cite{schmidhuber1992learning, katharopoulos2020transformers, schlag2021linear} trade performance for better computational efficiency. State space models \cite{gu2020hippo, gu2021efficiently, gu2023mamba} are popular alternatives that offer efficient, parallel training while keeping linear cost and efficient inference. The parallel training requirement forces only a linear recurrent relation between the timesteps. 
A common characteristic of such models is the relatively small, fixed memory that requires extreme compression. Despite recent progress, these models still underperform quadratic attention on many benchmarks \cite{arora2023zoology, jelassi2024repeat}.

Sparse attention methods aim to mitigate the quadratic cost of full attention by computing attention scores for only a subset of token pairs rather than the full attention matrix. These methods typically employ various heuristics to strategically identify which tokens and token relationships are the most important to process. This is often done by introducing special tokens that serve as higher-level representations of entire chunks of tokens, or by assuming emergent hierarchical structures within the attention patterns. For example, SepLLM~\cite{chen2024sepllm} uses separators in the sentence as special tokens that sparse attention focuses on. Sparse Transformer~\cite{child2019generating} uses static attention patterns to reduce computational complexity. Longformer~\cite{beltagy2020longformer} combines sliding window attention with additionally selected tokens globally available. BigBird~\cite{zaheer2020big} combines sliding window attention and global attention on selected tokens, while additionally including randomly selected tokens in the attention. Streaming LLM~\cite{xiao2024efficient} discovers and preserves attention sinks as a necessary component despite their inefficiency and combines them with sliding window attention. Some methods \cite{liu2023scissorhands, li2025snapkv, zefan2024pyramidkv} focus on post-training attention reduction, motivated by KV-cache reduction. Hash Attention\cite{desai2024hashattention} uses top-$k$ selection in the attention scores to induce sparsity and improve efficiency. However, learnable sparse attention that can also be used during training~\cite{yuan2025native} remains important as the quadratic cost of the self-attention mechanism is also problematic in the very costly pretaining phase. 

Mixture-of-Experts (MoE) \cite{shazeer2017outrageously} have emerged as a promising paradigm for scaling model capacity without a proportional increase in computational cost. By adaptively routing input tokens to specialized experts, MoE architectures selectively activate only a part of the network. MoEs applied to transformer feedforward networks \cite{lepikhin2020gshard, fedus2022switch} have been widely adapted in LLMs \cite{guo2025deepseek, jiang2024mixtral, shen2024jetmoe}. 

A crucial challenge in MoE is to learn a balanced routing, so that experts are utilized uniformly. Imbalanced routing leads to capacity bottlenecks when certain experts become overused while others are completely ignored. This phenomenon is called expert collapse \cite{shazeer2017outrageously}. Most approaches mitigate it by specific losses that penalize polarized expert selection \cite{lepikhin2020gshard}, while others propose alternative routing methods \cite{lewis2021base, roller2021hash}. Expert-Choice routing \cite{zhou2022mixture} inverts the selection problem, allowing each expert to choose its preferred tokens. This way, Expert-Choice routing achieves perfect load balancing by definition, at the cost that some tokens are ignored and some are overutilized. Expert-Choice routing, however, cannot be directly applied to autoregressive modeling as it uses a non-autoregressive top-$k$ operation over the tokens. MoD~\cite{raposo2024mixture} proposes methods to transfer nonautoregressive expert choice routing to an autoregressive model. We leave the investigation of their adaptation to MoSA for future work.

MoE is most often applied to the feedforward part of the transformer. In contrast, some works explore MoEs in the attention mechanism to reduce the high computational cost and memory.  Mixture-of-Attention Heads(MoA)~\cite{zhang2022moa} selects $k$ query transformations for each token and shares a single key and value projections similarly to Multi-Query Attention(MQA)~\cite{shazeer2019fast}. MoA allows for increasing the total number of query heads when using MQA without significantly increasing the computational cost. In contrast, MoSA selects tokens that are routed to full heads with separate queries, keys, and values (and consequently, outputs) utilizing perfect load balancing from expert choice routing for efficient sparse attention. This reduces the cost of each attention head significantly more than MoA and does not require MQA (although it might be combined for further benefits, which we leave for future work). Moreover, MoSA allows for KV-cache savings by reducing the number of selected keys, which is not possible with MoA, apart from the MQA benefit of having single KV transformations.
SwitchHead~\cite{csordas2023switchhead} reduces the number of heads (and therefore the number of computed attention matrices) by adding internal experts that can compensate for the lower number of heads. This is orthogonal to MoSA and possibly can be combined for further improvements. 
Multi-head attention as Mixture of Head Attention \cite{jin2024moh} proposes to use dynamic weights for the output projection in order to treat the heads as experts for tokens. However, it requires calculating all attention matrices, lacking the benefits of sparse computation.

Mixture-of-Depths(MoD)~\cite{raposo2024mixture} selects inputs to pass through a given entire transformer block to allow adaptive computation. This includes the attention mechanism. This produces efficiency gains in an FLOP-limited budget for the entire training. MoSA has multiple selection mechanisms, one for each head, and by increasing the number of heads it processes the sentence in a distributed way - each head processing its own chunk of the sentence.

\section{Limitations and Future work}\label{sec:future_work}

Due to the top-k selection over tokens, MoSA is non-autoregressive in nature and requires adaptations to be directly applicable to the autoregressive scenario. This is true not only for MoSA, but for all expert-choice routing methods, as well as for the Routing Transformer that uses non-autoregressive clustering. MoD proposed to solve this problem by learning an autoregressive classifier post-training to predict if the given token would have been selected by the non-autoregressive router or not. We consider exploring this issue in depth as an important future direction.

The perplexity gains do not always translate to downstream task performance (Section~\ref{sec:downstream}). This discrepancy stems from two distinct factors: First, sparse attention methods generally underperform on tasks consisting of short sequence lengths. Practitioners have shown that additional training with truncated sequences might alleviate this problem. Second, MoE architectures experience performance gaps in downstream tasks despite strong language modeling capabilities, although recent research demonstrates that instruction tuning can help significantly~\cite{shen2024mixture}. We consider exploring methods to mitigate the discrepancy between perplexity and downstream task performance in future work. 

Several promising research directions emerge from this work. Further exploration of MoSA's effectiveness on longer sequences remains an important direction. Furthermore, combining multiple sparse attention methods often leads to synergic improvements on long sequences~\cite{zaheer2020big, beltagy2020longformer}. Thus, we expect that combining other sparse head types with MoSA could lead to additional benefits. 

From an implementation perspective, developing specialized CUDA kernels would further improve efficiency. MoSA could be integrated with complementary approaches such as  MQA\cite{shazeer2019fast}, GQA\cite{ainslie2023gqa}, or SwitchHead\cite{csordas2023switchhead} to improve the efficiency even further. 

Furthermore, exploring MoSA on other modalities, particularly vision transformers, could yield valuable insights into the method's versatility across different data types and architectures.

\section{Conclusions}\label{sec:conclusions}

This paper introduces Mixture of Sparse Attention (MoSA), a novel attention architecture that selectively focuses on the most relevant tokens for the attention head, redirecting saved compute to create additional heads. MoSA reduces the computational complexity of attention from $O(T^2)$ to $O(k^2+T)$, where $T$ is the sequence length and $k$ is the number of selected tokens per head. 

Unlike other sparse attention methods that primarily show benefits for extremely long sequences, MoSA delivers substantial performance gains even in standard-length contexts. MoSA significantly outperforms both dense attention and sparse methods like fixed attention or the Routing Transformer, achieving up to 27\% perplexity improvement over dense baselines across models of different scales. We also demonstrated that MoSA can be used to reduce the resource requirements of the models, including a more than $50\%$ reduction in the KV-cache size. Additionally, our results indicate that MoSA maintains its superiority in long-sequence scenarios, outperforming other sparse attention methods in these contexts as well.

The efficiency and corresponding performance gains demonstrated by MoSA have significant implications for the design of adaptive architectures. MoSA or subsequent adaptive models stemming from MoSA can be used for reducing the training costs and environmental impact of large language models, potentially enabling more economical scaling while lowering energy consumption and carbon emissions. Given its versatility and performance advantages, we anticipate that MoSA will drive innovations in both transformer architecture research and industrial applications.

\section*{Acknowledgements}
For computer time, this research used Ibex managed by the Supercomputing Core
Laboratory at King Abdullah University of Science \& Technology (KAUST) in
Thuwal, Saudi Arabia.

The research reported in this publication was supported by funding from King Abdullah University of Science and Technology (KAUST) - Center of Excellence for Generative AI, under award number 5940.

\bibliography{mosa}

\begin{thebibliography}{70}
\providecommand{\natexlab}[1]{#1}
\providecommand{\url}[1]{\texttt{#1}}
\expandafter\ifx\csname urlstyle\endcsname\relax
  \providecommand{\doi}[1]{doi: #1}\else
  \providecommand{\doi}{doi: \begingroup \urlstyle{rm}\Url}\fi

\bibitem[Vaswani et~al.(2017)Vaswani, Shazeer, Parmar, Uszkoreit, Jones, Gomez,
  Kaiser, and Polosukhin]{vaswani2017attention}
Ashish Vaswani, Noam Shazeer, Niki Parmar, Jakob Uszkoreit, Llion Jones,
  Aidan~N. Gomez, Lukasz Kaiser, and Illia Polosukhin.
\newblock Attention is all you need.
\newblock In \emph{Proc. Advances in Neural Information Processing Systems
  (NIPS)}, pages 5998--6008, Long Beach, CA, USA, December 2017.

\bibitem[Brown et~al.(2020)Brown, Mann, Ryder, Subbiah, Kaplan, Dhariwal,
  Neelakantan, Shyam, Sastry, Askell, et~al.]{brown2020language}
Tom Brown, Benjamin Mann, Nick Ryder, Melanie Subbiah, Jared~D Kaplan, Prafulla
  Dhariwal, Arvind Neelakantan, Pranav Shyam, Girish Sastry, Amanda Askell,
  et~al.
\newblock Language models are few-shot learners.
\newblock In \emph{Proc. Advances in Neural Information Processing Systems
  (NeurIPS)}, pages 1877--1901, 2020.

\bibitem[Touvron et~al.(2023)Touvron, Lavril, Izacard, Martinet, Lachaux,
  Lacroix, Rozi{\`{e}}re, Goyal, Hambro, Azhar, Rodriguez, Joulin, Grave, and
  Lample]{touvron2023llama}
Hugo Touvron, Thibaut Lavril, Gautier Izacard, Xavier Martinet, Marie{-}Anne
  Lachaux, Timoth{\'{e}}e Lacroix, Baptiste Rozi{\`{e}}re, Naman Goyal, Eric
  Hambro, Faisal Azhar, Aur{\'{e}}lien Rodriguez, Armand Joulin, Edouard Grave,
  and Guillaume Lample.
\newblock {LLaMA}: Open and efficient foundation language models.
\newblock \emph{Preprint arXiv:2302.13971}, 2023.

\bibitem[Team et~al.(2024)Team, Georgiev, Lei, Burnell, Bai, Gulati, Tanzer,
  Vincent, Pan, Wang, et~al.]{team2024gemini}
Gemini Team, Petko Georgiev, Ving~Ian Lei, Ryan Burnell, Libin Bai, Anmol
  Gulati, Garrett Tanzer, Damien Vincent, Zhufeng Pan, Shibo Wang, et~al.
\newblock Gemini 1.5: Unlocking multimodal understanding across millions of
  tokens of context.
\newblock \emph{arXiv preprint arXiv:2403.05530}, 2024.

\bibitem[Grattafiori et~al.(2024)Grattafiori, Dubey, Jauhri, Pandey, Kadian,
  Al-Dahle, Letman, Mathur, Schelten, Vaughan, et~al.]{grattafiori2024llama}
Aaron Grattafiori, Abhimanyu Dubey, Abhinav Jauhri, Abhinav Pandey, Abhishek
  Kadian, Ahmad Al-Dahle, Aiesha Letman, Akhil Mathur, Alan Schelten, Alex
  Vaughan, et~al.
\newblock The llama 3 herd of models.
\newblock \emph{arXiv preprint arXiv:2407.21783}, 2024.

\bibitem[Gu et~al.(2020)Gu, Dao, Ermon, Rudra, and R{\'e}]{gu2020hippo}
Albert Gu, Tri Dao, Stefano Ermon, Atri Rudra, and Christopher R{\'e}.
\newblock Hippo: Recurrent memory with optimal polynomial projections.
\newblock In \emph{Proc. Advances in Neural Information Processing Systems
  (NeurIPS)}, volume~33, pages 1474--1487, 2020.

\bibitem[Gu et~al.(2022)Gu, Goel, and R{\'e}]{gu2021efficiently}
Albert Gu, Karan Goel, and Christopher R{\'e}.
\newblock Efficiently modeling long sequences with structured state spaces.
\newblock In \emph{Int. Conf. on Learning Representations (ICLR)}, 2022.

\bibitem[Gu and Dao(2023)]{gu2023mamba}
Albert Gu and Tri Dao.
\newblock Mamba: Linear-time sequence modeling with selective state spaces.
\newblock \emph{arXiv preprint arXiv:2312.00752}, 2023.

\bibitem[Wang et~al.(2024)Wang, Wang, Ding, Li, Wu, Rong, Kong, Huang, Li,
  Yang, et~al.]{wang2024state}
Xiao Wang, Shiao Wang, Yuhe Ding, Yuehang Li, Wentao Wu, Yao Rong, Weizhe Kong,
  Ju~Huang, Shihao Li, Haoxiang Yang, et~al.
\newblock State space model for new-generation network alternative to
  transformers: A survey.
\newblock \emph{arXiv preprint arXiv:2404.09516}, 2024.

\bibitem[Yang et~al.(2025)Yang, Kautz, and Hatamizadeh]{yang2025gated}
Songlin Yang, Jan Kautz, and Ali Hatamizadeh.
\newblock Gated delta networks: Improving mamba2 with delta rule.
\newblock In \emph{Int. Conf. on Learning Representations (ICLR)}, 2025.

\bibitem[Park et~al.(2024)Park, Park, Xiong, Lee, Cho, Oymak, Lee, and
  Papailiopoulos]{park2024mambaformer}
Jongho Park, Jaeseung Park, Zheyang Xiong, Nayoung Lee, Jaewoong Cho, Samet
  Oymak, Kangwook Lee, and Dimitris Papailiopoulos.
\newblock Can mamba learn how to learn? a comparative study on in-context
  learning tasks.
\newblock In \emph{Proc. Int. Conf. on Machine Learning (ICML)}, 2024.

\bibitem[Zuo et~al.(2022)Zuo, Liu, Jiao, Charles, Manavoglu, Zhao, and
  Gao]{zuo2022efficient}
Simiao Zuo, Xiaodong Liu, Jian Jiao, Denis Charles, Eren Manavoglu, Tuo Zhao,
  and Jianfeng Gao.
\newblock Efficient long sequence modeling via state space augmented
  transformer.
\newblock \emph{arXiv preprint arXiv:2212.08136}, 2022.

\bibitem[Lieber et~al.(2024)Lieber, Lenz, Bata, Cohen, Osin, Dalmedigos,
  Safahi, Meirom, Belinkov, Shalev-Shwartz, et~al.]{lieber2024jamba}
Opher Lieber, Barak Lenz, Hofit Bata, Gal Cohen, Jhonathan Osin, Itay
  Dalmedigos, Erez Safahi, Shaked Meirom, Yonatan Belinkov, Shai
  Shalev-Shwartz, et~al.
\newblock Jamba: A hybrid transformer-mamba language model.
\newblock \emph{arXiv preprint arXiv:2403.19887}, 2024.

\bibitem[Katharopoulos et~al.(2020)Katharopoulos, Vyas, Pappas, and
  Fleuret]{katharopoulos2020transformers}
Angelos Katharopoulos, Apoorv Vyas, Nikolaos Pappas, and Fran{\c{c}}ois
  Fleuret.
\newblock Transformers are {RNN}s: Fast autoregressive transformers with linear
  attention.
\newblock In \emph{Proc. Int. Conf. on Machine Learning (ICML)}, volume 119,
  pages 5156--5165, Virtual Only, 2020.

\bibitem[Schlag et~al.(2021)Schlag, Irie, and Schmidhuber]{schlag2021linear}
Imanol Schlag, Kazuki Irie, and J{\"{u}}rgen Schmidhuber.
\newblock Linear transformers are secretly fast weight programmers.
\newblock In \emph{Proc. Int. Conf. on Machine Learning (ICML)}, volume 139,
  pages 9355--9366, Virtual only, 2021.

\bibitem[Schmidhuber(1992)]{schmidhuber1992learning}
J{\"{u}}rgen Schmidhuber.
\newblock Learning to control fast-weight memories: An alternative to recurrent
  nets.
\newblock \emph{Neural Computation}, 4\penalty0 (1):\penalty0 131--139, 1992.

\bibitem[Qin et~al.(2022)Qin, Han, Sun, Li, Kong, Barnes, and
  Zhong]{qin2022-devil}
Zhen Qin, Xiaodong Han, Weixuan Sun, Dongxu Li, Lingpeng Kong, Nick Barnes, and
  Yiran Zhong.
\newblock The devil in linear transformer.
\newblock In \emph{Proc. Conf. on Empirical Methods in Natural Language
  Processing (EMNLP)}, pages 7025--7041, 2022.

\bibitem[Child et~al.(2019)Child, Gray, Radford, and
  Sutskever]{child2019generating}
Rewon Child, Scott Gray, Alec Radford, and Ilya Sutskever.
\newblock Generating long sequences with sparse transformers.
\newblock \emph{arXiv preprint arXiv:1904.10509}, 2019.

\bibitem[Zaheer et~al.(2020)Zaheer, Guruganesh, Dubey, Ainslie, Alberti,
  Onta\~non, Pham, Ravula, Wang, Yang, et~al.]{zaheer2020big}
Manzil Zaheer, Guru Guruganesh, Kumar~Avinava Dubey, Joshua Ainslie, Chris
  Alberti, Santiago Onta\~non, Philip Pham, Anirudh Ravula, Qifan Wang,
  Li~Yang, et~al.
\newblock Big bird: Transformers for longer sequences.
\newblock In \emph{Proc. Advances in Neural Information Processing Systems
  (NeurIPS)}, volume~33, pages 17283--17297, 2020.

\bibitem[Beltagy et~al.(2020)Beltagy, Peters, and Cohan]{beltagy2020longformer}
Iz~Beltagy, Matthew~E. Peters, and Arman Cohan.
\newblock Longformer: The long-document transformer.
\newblock \emph{arXiv:2004.05150}, 2020.

\bibitem[Arora et~al.(2024)Arora, Eyuboglu, Timalsina, Johnson, Poli, Zou,
  Rudra, and Ré]{arora2023zoology}
Simran Arora, Sabri Eyuboglu, Aman Timalsina, Isys Johnson, Michael Poli, James
  Zou, Atri Rudra, and Christopher Ré.
\newblock Zoology: Measuring and improving recall in efficient language models.
\newblock \emph{Int. Conf. on Learning Representations (ICLR)}, 2024.

\bibitem[Jelassi et~al.(2024)Jelassi, Brandfonbrener, Kakade, and
  Malach]{jelassi2024repeat}
Samy Jelassi, David Brandfonbrener, Sham~M Kakade, and Eran Malach.
\newblock Repeat after me: Transformers are better than state space models at
  copying.
\newblock In \emph{Proc. Int. Conf. on Machine Learning (ICML)}, 2024.

\bibitem[Tay et~al.(2021)Tay, Bahri, Metzler, Juan, Zhao, and
  Zheng]{tay2021synthesizer}
Yi~Tay, Dara Bahri, Donald Metzler, Da-Cheng Juan, Zhe Zhao, and Che Zheng.
\newblock Synthesizer: Rethinking self-attention for transformer models.
\newblock In \emph{Proc. Int. Conf. on Machine Learning (ICML)}, pages
  10183--10192, 2021.

\bibitem[Vyas et~al.(2020)Vyas, Katharopoulos, and Fleuret]{vyas2020fast}
Apoorv Vyas, Angelos Katharopoulos, and Fran{\c{c}}ois Fleuret.
\newblock Fast transformers with clustered attention.
\newblock In \emph{Proc. Advances in Neural Information Processing Systems
  (NeurIPS)}, volume~33, pages 21665--21674, 2020.

\bibitem[Roy et~al.(2021)Roy, Saffar, Vaswani, and Grangier]{roy2021efficient}
Aurko Roy, Mohammad Saffar, Ashish Vaswani, and David Grangier.
\newblock Efficient content-based sparse attention with routing transformers.
\newblock \emph{Transactions of the Association for Computational Linguistics
  (TACL)}, 9:\penalty0 53--68, 2021.

\bibitem[Bottou and Bengio(1994)]{bottou1994convergence}
Leon Bottou and Yoshua Bengio.
\newblock Convergence properties of the k-means algorithms.
\newblock In \emph{Proc. Advances in Neural Information Processing Systems
  (NIPS)}, volume~7, 1994.

\bibitem[Shazeer et~al.(2017)Shazeer, Mirhoseini, Maziarz, Davis, Le, Hinton,
  and Dean]{shazeer2017outrageously}
Noam Shazeer, Azalia Mirhoseini, Krzysztof Maziarz, Andy Davis, Quoc Le,
  Geoffrey Hinton, and Jeff Dean.
\newblock Outrageously large neural networks: The sparsely-gated
  mixture-of-experts layer.
\newblock In \emph{Int. Conf. on Learning Representations (ICLR)}, Toulon,
  France, April 2017.

\bibitem[Fedus et~al.(2022)Fedus, Zoph, and Shazeer]{fedus2022switch}
William Fedus, Barret Zoph, and Noam Shazeer.
\newblock Switch transformers: Scaling to trillion parameter models with simple
  and efficient sparsity.
\newblock \emph{Journal of Machine Learning Research (JMLR)}, 23\penalty0
  (1):\penalty0 5232--5270, 2022.

\bibitem[Zhou et~al.(2022)Zhou, Lei, Liu, Du, Huang, Zhao, Dai, Le, Laudon,
  et~al.]{zhou2022mixture}
Yanqi Zhou, Tao Lei, Hanxiao Liu, Nan Du, Yanping Huang, Vincent Zhao, Andrew~M
  Dai, Quoc~V Le, James Laudon, et~al.
\newblock Mixture-of-experts with expert choice routing.
\newblock In \emph{Proc. Advances in Neural Information Processing Systems
  (NeurIPS)}, volume~35, pages 7103--7114, 2022.

\bibitem[Zhang et~al.(2022)Zhang, Shen, Huang, Zhou, Rong, and
  Xiong]{zhang2022moa}
Xiaofeng Zhang, Yikang Shen, Zeyu Huang, Jie Zhou, Wenge Rong, and Zhang Xiong.
\newblock Mixture of attention heads: Selecting attention heads per token.
\newblock In \emph{Proc. Conf. on Empirical Methods in Natural Language
  Processing (EMNLP)}, pages 4150--4162, Abu Dhabi, United Arab Emirates,
  December 2022.

\bibitem[Csord\'as et~al.(2024)Csord\'as, Pi\k{e}kos, Irie, and
  Schmidhuber]{csordas2023switchhead}
R\'obert Csord\'as, Piotr Pi\k{e}kos, Kazuki Irie, and J\"urgen Schmidhuber.
\newblock Switchhead: Accelerating transformers with mixture-of-experts
  attention.
\newblock In \emph{Proc. Advances in Neural Information Processing Systems
  (NeurIPS)}, Vancouver, Canada, December 2024.

\bibitem[Liu et~al.(2023)Liu, Desai, Liao, Wang, Xie, Xu, Kyrillidis, and
  Shrivastava]{liu2023scissorhands}
Zichang Liu, Aditya Desai, Fangshuo Liao, Weitao Wang, Victor Xie, Zhaozhuo Xu,
  Anastasios Kyrillidis, and Anshumali Shrivastava.
\newblock Scissorhands: Exploiting the persistence of importance hypothesis for
  llm kv cache compression at test time.
\newblock In \emph{Proc. Advances in Neural Information Processing Systems
  (NeurIPS)}, volume~36, pages 52342--52364, 2023.

\bibitem[Li et~al.(2025)Li, Huang, Yang, Venkitesh, Locatelli, Ye, Cai, Lewis,
  and Chen]{li2025snapkv}
Yuhong Li, Yingbing Huang, Bowen Yang, Bharat Venkitesh, Acyr Locatelli,
  Hanchen Ye, Tianle Cai, Patrick Lewis, and Deming Chen.
\newblock Snapkv: Llm knows what you are looking for before generation.
\newblock In \emph{Proc. Advances in Neural Information Processing Systems
  (NeurIPS)}, volume~37, pages 22947--22970, 2025.

\bibitem[Cai et~al.(2024)Cai, Zhang, Gao, Liu, Liu, Lu, Xiong, Dong, Chang, Hu,
  and Wen]{zefan2024pyramidkv}
Zefan Cai, Yichi Zhang, Bofei Gao, Yuliang Liu, Tianyu Liu, Keming Lu, Wayne
  Xiong, Yue Dong, Baobao Chang, Junjie Hu, and Xiao Wen.
\newblock Pyramidkv: Dynamic kv cache compression based on pyramidal
  information funneling.
\newblock \emph{arXiv preprint arXiv:2406.02069}, 2024.

\bibitem[Lepikhin et~al.(2021)Lepikhin, Lee, Xu, Chen, Firat, Huang, Krikun,
  Shazeer, and Chen]{lepikhin2020gshard}
Dmitry Lepikhin, HyoukJoong Lee, Yuanzhong Xu, Dehao Chen, Orhan Firat, Yanping
  Huang, Maxim Krikun, Noam Shazeer, and Zhifeng Chen.
\newblock Gshard: Scaling giant models with conditional computation and
  automatic sharding.
\newblock In \emph{Int. Conf. on Learning Representations (ICLR)}, 2021.

\bibitem[Shazeer(2019)]{shazeer2019fast}
Noam Shazeer.
\newblock Fast transformer decoding: One write-head is all you need.
\newblock \emph{arXiv preprint arXiv:1911.02150}, 2019.

\bibitem[Yuan et~al.(2025)Yuan, Gao, Dai, Luo, Zhao, Zhang, Xie, Wei, Wang,
  Xiao, et~al.]{yuan2025native}
Jingyang Yuan, Huazuo Gao, Damai Dai, Junyu Luo, Liang Zhao, Zhengyan Zhang,
  Zhenda Xie, YX~Wei, Lean Wang, Zhiping Xiao, et~al.
\newblock Native sparse attention: Hardware-aligned and natively trainable
  sparse attention.
\newblock \emph{arXiv preprint arXiv:2502.11089}, 2025.

\bibitem[Csord\'as et~al.(2023)Csord\'as, Irie, and
  Schmidhuber]{csordas2023approximating}
R\'obert Csord\'as, Kazuki Irie, and J\"urgen Schmidhuber.
\newblock Approximating two-layer feedforward networks for efficient
  transformers.
\newblock In \emph{Findings of the Association for Computational Linguistics:
  {EMNLP} 2023}, November 2023.

\bibitem[Dao et~al.(2022)Dao, Fu, Ermon, Rudra, and R{\'{e}}]{dao2022flash}
Tri Dao, Daniel~Y. Fu, Stefano Ermon, Atri Rudra, and Christopher R{\'{e}}.
\newblock Flash{A}ttention: Fast and memory-efficient exact attention with
  {IO}-awareness.
\newblock In \emph{Proc. Advances in Neural Information Processing Systems
  (NeurIPS)}, New Orleans, Louisiana, USA, December 2022.

\bibitem[Paszke et~al.(2019)Paszke, Gross, Massa, Lerer, Bradbury, Chanan,
  Killeen, Lin, Gimelshein, Antiga, Desmaison, Kopf, Yang, DeVito, Raison,
  Tejani, Chilamkurthy, Steiner, Fang, Bai, and Chintala]{paszke2019pytorch}
Adam Paszke, Sam Gross, Francisco Massa, Adam Lerer, James Bradbury, Gregory
  Chanan, Trevor Killeen, Zeming Lin, Natalia Gimelshein, Luca Antiga, Alban
  Desmaison, Andreas Kopf, Edward Yang, Zachary DeVito, Martin Raison, Alykhan
  Tejani, Sasank Chilamkurthy, Benoit Steiner, Lu~Fang, Junjie Bai, and Soumith
  Chintala.
\newblock {PyTorch}: An imperative style, high-performance deep learning
  library.
\newblock In \emph{Proc. Advances in Neural Information Processing Systems
  (NeurIPS)}, pages 8024--8035, Vancouver, Canada, December 2019.

\bibitem[Su et~al.(2021)Su, Lu, Pan, Wen, and Liu]{su2021roformer}
Jianlin Su, Yu~Lu, Shengfeng Pan, Bo~Wen, and Yunfeng Liu.
\newblock {RoFormer}: Enhanced transformer with rotary position embedding.
\newblock \emph{Preprint arXiv:2104.09864}, 2021.

\bibitem[Kudo and Richardson(2018)]{KudoR18}
Taku Kudo and John Richardson.
\newblock Sentencepiece: {A} simple and language independent subword tokenizer
  and detokenizer for neural text processing.
\newblock In \emph{Proc. Conf. on Empirical Methods in Natural Language
  Processing (EMNLP)}, pages 66--71, Brussels, Belgium, October 2018.

\bibitem[Sennrich et~al.(2016)Sennrich, Haddow, and Birch]{sennrich16bpe}
Rico Sennrich, Barry Haddow, and Alexandra Birch.
\newblock Neural machine translation of rare words with subword units.
\newblock In \emph{Proc. Association for Computational Linguistics (ACL)},
  pages 1715--1725, Berlin, Germany, August 2016.

\bibitem[Schuster and Nakajima(2012)]{SchusterN12}
Mike Schuster and Kaisuke Nakajima.
\newblock Japanese and korean voice search.
\newblock In \emph{Proc. {IEEE} Int. Conf. on Acoustics, Speech and Signal
  Processing (ICASSP)}, pages 5149--5152, Kyoto, Japan, March 2012.

\bibitem[Raffel et~al.(2020)Raffel, Shazeer, Roberts, Lee, Narang, Matena,
  Zhou, Li, and Liu]{raffel2020exploring}
Colin Raffel, Noam Shazeer, Adam Roberts, Katherine Lee, Sharan Narang, Michael
  Matena, Yanqi Zhou, Wei Li, and Peter~J. Liu.
\newblock Exploring the limits of transfer learning with a unified text-to-text
  transformer.
\newblock \emph{Journal of Machine Learning Research (JMLR)}, 21:\penalty0
  140:1--140:67, 2020.

\bibitem[Kingma and Ba(2015)]{kingma2014adam}
Diederik~P. Kingma and Jimmy Ba.
\newblock Adam: {A} method for stochastic optimization.
\newblock In Yoshua Bengio and Yann LeCun, editors, \emph{Int. Conf. on
  Learning Representations (ICLR)}, San Diego, CA, USA, May 2015.

\bibitem[Zhang et~al.(2023)Zhang, Ram, Hawkins, Zha, and
  Zhao]{zhang2023efficient}
Qingru Zhang, Dhananjay Ram, Cole Hawkins, Sheng Zha, and Tuo Zhao.
\newblock Efficient long-range transformers: You need to attend more, but not
  necessarily at every layer.
\newblock In \emph{Proc. Conf. on Empirical Methods in Natural Language
  Processing (EMNLP)}, 2023.

\bibitem[Xiao et~al.(2024)Xiao, Tian, Chen, Han, and Lewis]{xiao2024efficient}
Guangxuan Xiao, Yuandong Tian, Beidi Chen, Song Han, and Mike Lewis.
\newblock Efficient streaming language models with attention sinks.
\newblock In \emph{Int. Conf. on Learning Representations (ICLR)}, 2024.

\bibitem[Kitaev et~al.(2020)Kitaev, {\L}ukasz, and
  Levskaya]{kitaev2020reformer}
Nikita Kitaev, Kaiser {\L}ukasz, and Anselm Levskaya.
\newblock Reformer: The efficient transformer.
\newblock In \emph{Int. Conf. on Learning Representations (ICLR)}, 2020.

\bibitem[Xu et~al.(2025)Xu, Ping, Xu, Liu, Wang, Shoeybi, Li, and
  Catanzaro]{xu2025128k}
Chejian Xu, Wei Ping, Peng Xu, Zihan Liu, Boxin Wang, Mohammad Shoeybi, Bo~Li,
  and Bryan Catanzaro.
\newblock From 128k to 4m: Efficient training of ultra-long context large
  language models.
\newblock \emph{arXiv preprint arXiv:2504.06214}, 2025.

\bibitem[Paperno et~al.(2016)Paperno, Kruszewski, Lazaridou, Pham, Bernardi,
  Pezzelle, Baroni, Boleda, and Fern{\'{a}}ndez]{paperno2016lambada}
Denis Paperno, Germ{\'{a}}n Kruszewski, Angeliki Lazaridou, Quan~Ngoc Pham,
  Raffaella Bernardi, Sandro Pezzelle, Marco Baroni, Gemma Boleda, and Raquel
  Fern{\'{a}}ndez.
\newblock The {LAMBADA} dataset: Word prediction requiring a broad discourse
  context.
\newblock In \emph{Proc. Association for Computational Linguistics (ACL)},
  Berlin, Germany, August 2016.

\bibitem[Sakaguchi et~al.(2020)Sakaguchi, Bras, Bhagavatula, and
  Choi]{sakaguchi2020winogrande}
Keisuke Sakaguchi, Ronan~Le Bras, Chandra Bhagavatula, and Yejin Choi.
\newblock Winogrande: An adversarial winograd schema challenge at scale.
\newblock In \emph{Proc. {AAAI} Conf. on Artificial Intelligence}, pages
  8732--8740, New York, NY, USA, February 2020.

\bibitem[Warstadt et~al.(2020)Warstadt, Parrish, Liu, Mohananey, Peng, Wang,
  and Bowman]{warstadt2020blimp}
Alex Warstadt, Alicia Parrish, Haokun Liu, Anhad Mohananey, Wei Peng,
  Sheng{-}Fu Wang, and Samuel~R. Bowman.
\newblock {BLiMP}: The benchmark of linguistic minimal pairs for {E}nglish.
\newblock \emph{Transactions of the Association for Computational Linguistics
  (TACL)}, 8:\penalty0 377--392, 2020.

\bibitem[Zellers et~al.(2019)Zellers, Holtzman, Bisk, Farhadi, and
  Choi]{zellers2019hellaswag}
Rowan Zellers, Ari Holtzman, Yonatan Bisk, Ali Farhadi, and Yejin Choi.
\newblock Hellaswag: Can a machine really finish your sentence?
\newblock In \emph{Proc. Association for Computational Linguistics (ACL)},
  pages 4791--4800, Florence, Italy, August 2019.

\bibitem[Bisk et~al.(2020)Bisk, Zellers, Bras, Gao, and Choi]{bisk2020piqa}
Yonatan Bisk, Rowan Zellers, Ronan~Le Bras, Jianfeng Gao, and Yejin Choi.
\newblock {PIQA:} reasoning about physical commonsense in natural language.
\newblock In \emph{Proc. {AAAI} Conf. on Artificial Intelligence}, pages
  7432--7439, New York, NY, USA, February 2020. {AAAI} Press.

\bibitem[Clark et~al.(2018)Clark, Cowhey, Etzioni, Khot, Sabharwal, Schoenick,
  and Tafjord]{clark2018think}
Peter Clark, Isaac Cowhey, Oren Etzioni, Tushar Khot, Ashish Sabharwal, Carissa
  Schoenick, and Oyvind Tafjord.
\newblock Think you have solved question answering? try {ARC}, the {AI2}
  reasoning challenge.
\newblock \emph{Preprint arXiv:1803.05457}, 2018.

\bibitem[Zoph et~al.(2022)Zoph, Bello, Kumar, Du, Huang, Dean, Shazeer, and
  Fedus]{zoph2022st}
Barret Zoph, Irwan Bello, Sameer Kumar, Nan Du, Yanping Huang, Jeff Dean, Noam
  Shazeer, and William Fedus.
\newblock St-moe: Designing stable and transferable sparse expert models.
\newblock \emph{arXiv preprint arXiv:2202.08906}, 2022.

\bibitem[Shen et~al.(2024{\natexlab{a}})Shen, Hou, Zhou, Du, Longpre, Wei,
  Chung, Zoph, Fedus, Chen, et~al.]{shen2024mixture}
Sheng Shen, Le~Hou, Yanqi Zhou, Nan Du, Shayne Longpre, Jason Wei, Hyung~Won
  Chung, Barret Zoph, William Fedus, Xinyun Chen, et~al.
\newblock Mixture-of-experts meets instruction tuning: A winning combination
  for large language models.
\newblock In \emph{Int. Conf. on Learning Representations (ICLR)},
  2024{\natexlab{a}}.

\bibitem[Choromanski et~al.(2021)Choromanski, Likhosherstov, Dohan, Song, Gane,
  Sarl{\'{o}}s, Hawkins, Davis, Mohiuddin, Kaiser, Belanger, Colwell, and
  Weller]{choromanski2021rethinking}
Krzysztof~Marcin Choromanski, Valerii Likhosherstov, David Dohan, Xingyou Song,
  Andreea Gane, Tam{\'{a}}s Sarl{\'{o}}s, Peter Hawkins, Jared~Quincy Davis,
  Afroz Mohiuddin, Lukasz Kaiser, David~Benjamin Belanger, Lucy~J. Colwell, and
  Adrian Weller.
\newblock Rethinking attention with performers.
\newblock In \emph{Int. Conf. on Learning Representations (ICLR)}, Virtual
  only, May 2021.

\bibitem[Chen et~al.(2024)Chen, Shi, Li, Gao, Ren, Chen, Jiang, Li, Liu, and
  Huang]{chen2024sepllm}
Guoxuan Chen, Han Shi, Jiawei Li, Yihang Gao, Xiaozhe Ren, Yimeng Chen, Xin
  Jiang, Zhenguo Li, Weiyang Liu, and Chao Huang.
\newblock Sepllm: Accelerate large language models by compressing one segment
  into one separator.
\newblock \emph{arXiv preprint arXiv:2412.12094}, 2024.

\bibitem[Desai et~al.(2024)Desai, Yang, Cuadron, Klimovic, Zaharia, Gonzalez,
  and Stoica]{desai2024hashattention}
Aditya Desai, Shuo Yang, Alejandro Cuadron, Ana Klimovic, Matei Zaharia,
  Joseph~E Gonzalez, and Ion Stoica.
\newblock Hashattention: Semantic sparsity for faster inference.
\newblock \emph{arXiv preprint arXiv:2412.14468}, 2024.

\bibitem[Guo et~al.(2025)Guo, Yang, Zhang, Song, Zhang, Xu, Zhu, Ma, Wang, Bi,
  et~al.]{guo2025deepseek}
Daya Guo, Dejian Yang, Haowei Zhang, Junxiao Song, Ruoyu Zhang, Runxin Xu,
  Qihao Zhu, Shirong Ma, Peiyi Wang, Xiao Bi, et~al.
\newblock Deepseek-r1: Incentivizing reasoning capability in llms via
  reinforcement learning.
\newblock \emph{arXiv preprint arXiv:2501.12948}, 2025.

\bibitem[Jiang et~al.(2024)Jiang, Sablayrolles, Roux, Mensch, Savary, Bamford,
  Chaplot, Casas, Hanna, Bressand, et~al.]{jiang2024mixtral}
Albert~Q Jiang, Alexandre Sablayrolles, Antoine Roux, Arthur Mensch, Blanche
  Savary, Chris Bamford, Devendra~Singh Chaplot, Diego de~las Casas, Emma~Bou
  Hanna, Florian Bressand, et~al.
\newblock Mixtral of experts.
\newblock \emph{arXiv preprint arXiv:2401.04088}, 2024.

\bibitem[Shen et~al.(2024{\natexlab{b}})Shen, Guo, Cai, and
  Qin]{shen2024jetmoe}
Yikang Shen, Zhen Guo, Tianle Cai, and Zengyi Qin.
\newblock Jetmoe: Reaching llama2 performance with 0.1 m dollars.
\newblock \emph{arXiv preprint arXiv:2404.07413}, 2024{\natexlab{b}}.

\bibitem[Lewis et~al.(2021)Lewis, Bhosale, Dettmers, Goyal, and
  Zettlemoyer]{lewis2021base}
Mike Lewis, Shruti Bhosale, Tim Dettmers, Naman Goyal, and Luke Zettlemoyer.
\newblock {BASE} layers: Simplifying training of large, sparse models.
\newblock In Marina Meila and Tong Zhang, editors, \emph{Proc. Int. Conf. on
  Machine Learning (ICML)}, volume 139, pages 6265--6274, Virtual only, July
  2021.

\bibitem[Roller et~al.(2021)Roller, Sukhbaatar, Weston, et~al.]{roller2021hash}
Stephen Roller, Sainbayar Sukhbaatar, Jason Weston, et~al.
\newblock Hash layers for large sparse models.
\newblock In \emph{Proc. Advances in Neural Information Processing Systems
  (NeurIPS)}, volume~34, pages 17555--17566, 2021.

\bibitem[Raposo et~al.(2024)Raposo, Ritter, Richards, Lillicrap, Humphreys, and
  Santoro]{raposo2024mixture}
David Raposo, Sam Ritter, Blake Richards, Timothy Lillicrap, Peter~Conway
  Humphreys, and Adam Santoro.
\newblock Mixture-of-depths: Dynamically allocating compute in
  transformer-based language models.
\newblock \emph{arXiv preprint arXiv:2404.02258}, 2024.

\bibitem[Jin et~al.(2024)Jin, Zhu, Yuan, and Yan]{jin2024moh}
Peng Jin, Bo~Zhu, Li~Yuan, and Shuicheng Yan.
\newblock Moh: Multi-head attention as mixture-of-head attention.
\newblock \emph{arXiv preprint arXiv:2410.11842}, 2024.

\bibitem[Ainslie et~al.(2023)Ainslie, Lee-Thorp, de~Jong, Zemlyanskiy,
  Lebr{\'o}n, and Sanghai]{ainslie2023gqa}
Joshua Ainslie, James Lee-Thorp, Michiel de~Jong, Yury Zemlyanskiy, Federico
  Lebr{\'o}n, and Sumit Sanghai.
\newblock Gqa: Training generalized multi-query transformer models from
  multi-head checkpoints.
\newblock In \emph{Proc. Conf. on Empirical Methods in Natural Language
  Processing (EMNLP)}, pages 4895--4901, 2023.

\bibitem[Xiong et~al.(2020)Xiong, Yang, He, Zheng, Zheng, Xing, Zhang, Lan,
  Wang, and Liu]{xiong2020layer}
Ruibin Xiong, Yunchang Yang, Di~He, Kai Zheng, Shuxin Zheng, Chen Xing,
  Huishuai Zhang, Yanyan Lan, Liwei Wang, and Tie{-}Yan Liu.
\newblock On layer normalization in the transformer architecture.
\newblock In \emph{Proc. Int. Conf. on Machine Learning (ICML)}, volume 119,
  pages 10524--10533, Virtual Only, July 2020.

\end{thebibliography}
\bibliographystyle{unsrtnat}

\appendix

\section{FLOPs cost derivation}\label{app:FLOPs}
In this section, we derive the FLOP cost for dense and MoSA heads, and the total FLOPs necessary for the forward pass of each model. 

Multiplying matrices of shape $[i, j]$ and $[j, k]$ takes precisely $(2j-1)ik$ FLOPs. For simplicity, following common practice, we approximate it by $2jik$.

In the dense attention layer, calculating each projection (e.g., $Q_i = xW_{Q_i}$) requires $2hh'T$ FLOPs. Computing the attention matrix $QK^\top$, and multiplying the attention matrix by values $V$ both cost $2h'T^2$ FLOPs.

Calculating the projections and attention in the MoSA head is identical, except that now we are operating on $k$ tokens instead of $T$. The MoSA head involves an additional routing overhead. Calculating the routing scores costs $2hT$ FLOPs, and multiplying the intermediate values in the matrix $\in \mathbb{R}^{k \times h'}$ by the scores costs an additional $h'k$ FLOPs per head.

The cost of a single head of the dense and MoSA heads are:
\begin{align*}
\text{FLOP}_{\text{dense}} &= \underbrace{\textcolor{blue}{8hh'T}}_{\text{Q,K,V,O mappings}} + \underbrace{\textcolor{orange}{4h'T^2}}_{\text{Attention}} \\[10pt]
\text{FLOP}_{\text{mosa}} &= \underbrace{\textcolor{blue}{8hh'k}}_{\text{Q,K,V,O mappings}} + \underbrace{\textcolor{orange}{4h'k^2}}_{\text{Attention}} + \underbrace{\textcolor{purple}{2hT + h'k}}_{\text{routing overhead}}
\end{align*}

For the multihead version, the FLOPs are multiplied by the number of heads $H$. There is an additional cost caused by summing the head contributions to a single output (Equations \ref{eq:mha} and \ref{eq:mosa_mha}). However, this is already taken into account by the $2hh'TH$ cost of the output projection for multiple heads: $
    H(2h'-1)hT + (H-1)hT = (2h'H-1)hT \approx 2hh'TH$.

Note that in the standard notation~\cite{vaswani2017attention}, the heads are first concatenated and then transformed with a single output projection instead of splitting the output operation into individual head transformations and summing. 
However, the result and the derivation of the FLOP counts are the same. 

In the feedforward block, the intermediate layer has a typical size of $4h$. Therefore, the cost of the block is equal to $16h^2T$. Therefore, the FLOP cost of the forward pass of the entire model with $l$ layers, a hybrid attention with $H_{dense}$ dense heads and $H_{mosa}$ MoSA heads is equal to:
\[ lH_{dense}(8hh'T+4h'T^2) + lH_{mosa}(8hh'k+4h'k^2+2hT+h'k) + 16lh^2T\]
We omit the operations related to layer normalizations, residuals, and token embeddings from the FLOP calculations as they are negligible compared to the rest and represent an identical overhead for both dense and MoSA models. Thus, incorporating them does not influence the FLOP-matching process. This is also true for the feedforward block; yet, we still included it because it constitutes a significant portion of the total cost.
We present the FLOP cost of all of our model classes (\textit{Tiny, Small, Medium} and \textit{Large}) in Table~\ref{tab:model_hyperparameters}.

All models are based on the transformer architecture with Pre-layer normalisation\cite{xiong2020layer}. Each model class \textit{Tiny, Small, Medium} and \textit{Large} follows the hyperparameters of the dense model. The necessary forward pass FLOPs are calculated according to Sec.~\ref{sec:flop_calculation}. The number of heads in the sparse models is set so that the resulting model is FLOP-matched to the dense baseline as closely as possible. When this is not perfectly possible, we ensure that its FLOP count never exceeds that of the baseline. For pure MoSA, all heads are replaced with MoSA heads. For the hybrid sparse models, 4 dense heads are kept, and the remaining ones are replaced with sparse heads. 

\section{Analysing Hybrid Models}\label{app:hybrid}

\begin{figure}
    \centering
    \includegraphics[width=0.5\linewidth]{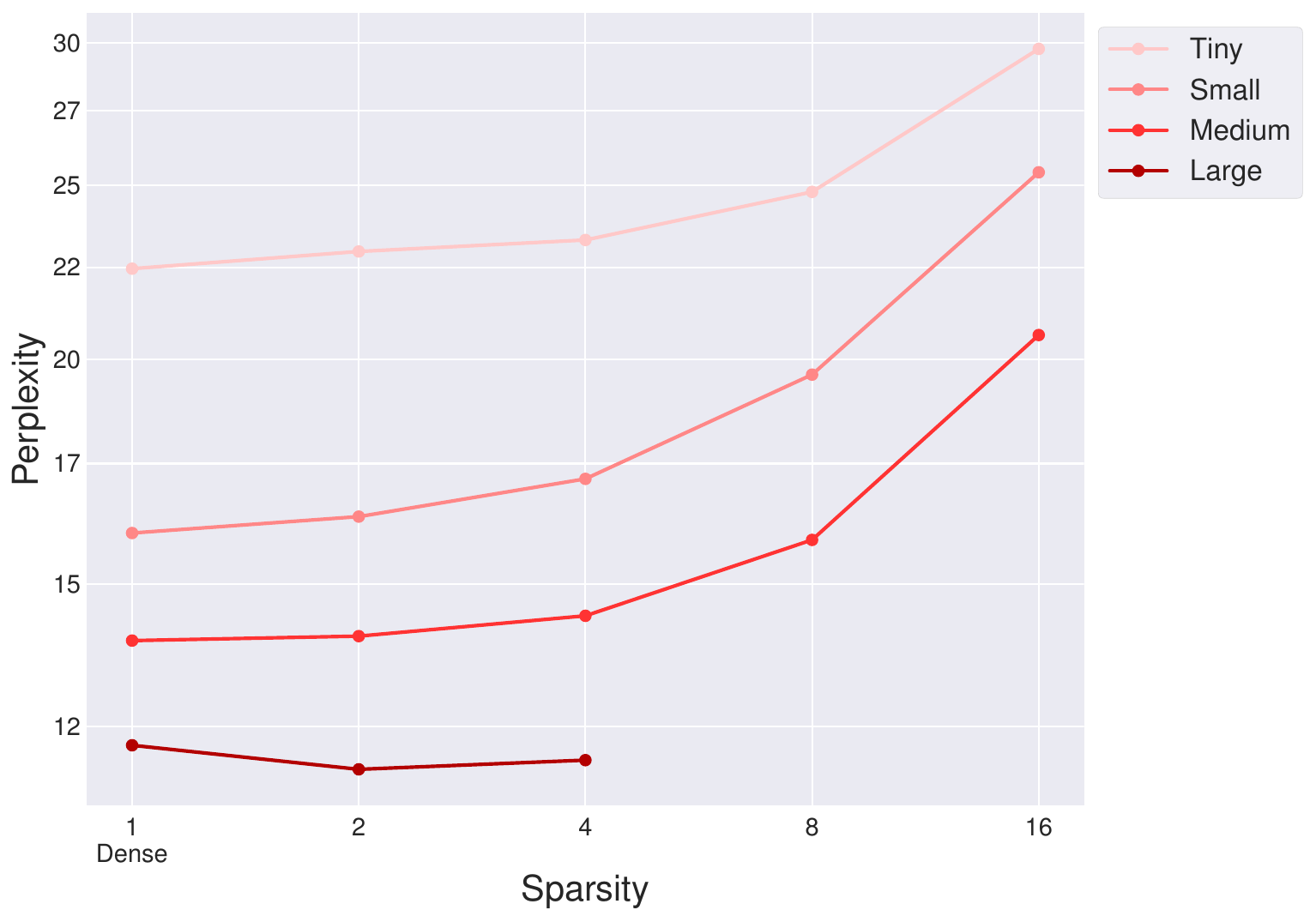}
    \caption{Perplexity of IsoFLOP matching models under pure MoSA setting. Each curve corresponds to a given FLOP budget. For a given sparsity, we replace all dense heads with a FLOP equivalent number of MoSA heads. In contrast to Fig~\ref{fig:main_results}, sparse models fail to outperform the baseline (apart from the Large model). This demonstrates the symbiotic relation between dense heads and MoSA heads in the hybrid model.}
    \label{fig:isoflop_0shared}
\end{figure}

While the learned sparse attention can theoretically capture any attention pattern, the introduction of the routing mechanism complicates the learning dynamics. The router and the attention weights must be learned jointly. The router needs to identify relevant token pairs, while the attention weights learn to process these selected interactions. This interdependence can lead to training instabilities, particularly in the early stages, when router decisions are largely random. Poor initial routing can prevent attention heads from learning meaningful patterns, while the lack of meaningful patterns prevents the router from learning to select important tokens, creating a vicious circle.

Our preliminary experiments have shown that pure MoSA models without additional dense heads fail to improve the perplexity of dense baselines. To verify this, we conducted a study similar to our main results in Sec. \ref{sec:isoflop}. We gradually increase the sparsity by replacing all dense heads with MoSA heads while maintaining an identical FLOP count to the baseline. We do this by finding the maximum number of MoSA heads for which the FLOP count remains lower than the baseline. The results, shown in Fig.~\ref{fig:isoflop_0shared}, demonstrate that increasing sparsity monotonically worsens model performance in most settings. This performance degradation with pure MoSA heads likely stems from the stability issues explained in the previous paragraph.

Interestingly, the largest model is an exception, and initially there is a visible improvement from 12.20 baseline perplexity to 11.83 perplexity of the FLOP-matched pure MoSA model with sparsity 2. This is still significantly worse than the 10.58 perplexity of the hybrid model with sparsity 4. Moreover, the saturation is much faster than for hybrid models. For hybrid models, the sparsity around 32 or 64 seems to be optimal. In contrast, for the MoSA-only model, the best perplexity is reached for sparsity 2 for the \textit{Large} budget and 1 for the smaller ones. However, the conclusion is consistent across all scales: hybrid MoSA models significantly outperform MoSA-only models, which generally underperform the dense baseline. Thus, hybridization seems necessary.

\begin{figure}
    \centering
    \includegraphics[width=0.7\linewidth]{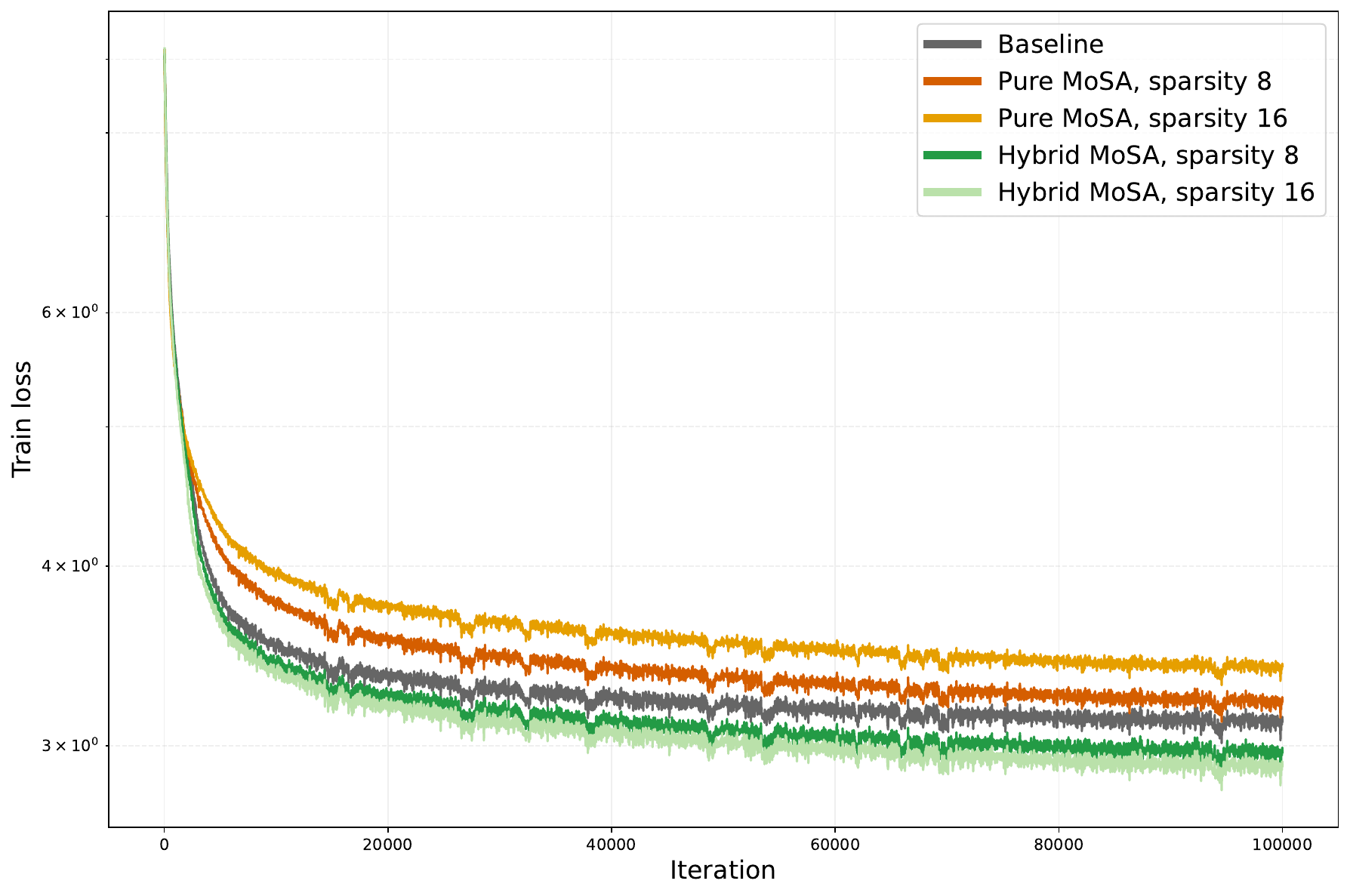}
    \caption{Training losses of the Tiny models comparing the baseline, pure MoSA, and hybrid models. The dense baseline clearly divides the models into two groups: all pure MoSA models perform worse (higher loss), while all hybrid models demonstrate superior performance (lower loss). Notably, increasing sparsity intensifies the difference for both model types: hybrid models achieve progressively lower loss with greater sparsity, whereas pure MoSA models show increasingly higher loss as sparsity increases. Additionally, the early training phase (between 5,000 and 10,000 steps) reveals a distinct pattern where pure MoSA models experience a more rapid slowdown in their learning progress compared to both dense and hybrid models.  }
    \label{fig:learning_curves}
\end{figure}

The impact of sparsification is also visible in the training characteristics. Compared to the baseline, pure MoSA models start to plateau faster. While the losses of dense and hybrid models continue to show steep initial improvement, pure MoSA models slow down much sooner. This supports our hypothesis about the difficulty of learning the routing and attention simultaneously. We compare the training losses in Fig.~\ref{fig:learning_curves}.

\paragraph{Optimal Number of Dense Heads}

Hybrid models consistently outperform pure MoSA models. This raises a natural question: What is the optimal ratio of dense to sparse heads and how does this ratio relate to the sparsity rate? 

\begin{figure}
    \centering
    \includegraphics[width=0.5\linewidth]{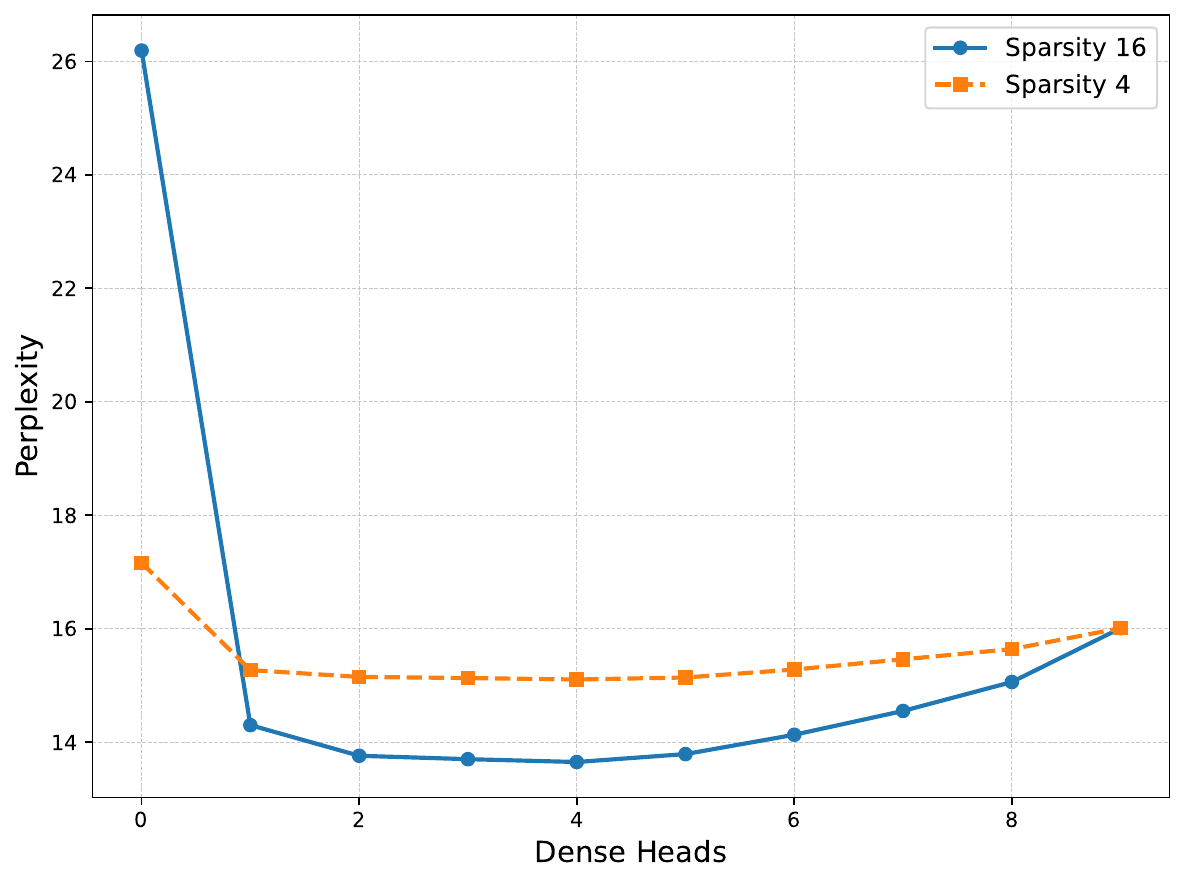}
    \caption{Perplexity of the FLOP matched models with a different number of dense heads for sparsities 4 and 16. 9 dense heads correspond to the dense baseline. }
    \label{fig:dense_heads}
\end{figure}

To answer these questions, we conducted a series of experiments in which we varied both the sparsity factor of MoSA heads and the number of dense heads while keeping the total FLOP budget constant. We choose to use the \textit{small} model and investigate sparsities $\rho=4$ and $\rho=16$, while we set the number of dense heads in the hybrid model from $0$ to $9$ (full dense model) and adapt the number of sparse heads to match the FLOP budget. Our results are shown in Fig.~\ref{fig:dense_heads}.
We can see that the optimal number of dense heads in this case is 4 and is sparsity-agnostic. Because of this, we chose to use 4 dense heads in our main experiments in Sect.~\ref{sec:isoflop}. Furthermore, we observe that it is critical to have at least one dense head. Having more than one has diminishing returns, and having more than 4 has a negative effect on the performance. The plot also shows that the lack of dense heads is more hurtful for models with higher sparsities. We conclude that in our case, 4 heads are sufficient to stabilize the training, and it is better to allocate the remaining FLOP budget to the more efficient MoSA heads.

\renewcommand{\arraystretch}{1.2}

\section{Details of the Models}\label{app:models_details}

In the Table~\ref{tab:model_hyperparameters} we list hyperparameters all of dense baselines.
\begin{table}[htbp]
    \centering
    \renewcommand{\arraystretch}{1.2}
    \begin{tabular}{l|rrrr}
        \toprule
         & \textbf{Tiny} & \textbf{Small} & \textbf{Medium} & \textbf{Large} \\
        \midrule
        FLOPs per pass (G) & 54.76 & 219.85 & 430.70 & 1,130.65 \\
        Layers & 6 & 9 & 18 & 27 \\
        Hidden size & 512 & 1,024 & 1,024 & 1,280 \\
        Feedforward hidden size & 2,048 & 4,096 & 4,096 & 5,120 \\
        Head hidden size & 64 & 64 & 64 & 64 \\
        Number of heads & 9 & 9 & 9 & 16 \\
        \bottomrule
    \end{tabular}
    \caption{Hyperparameters of the different model variants and the corresponding FLOP cost of the forward pass for a sequence length of $T=1024$.}
    \label{tab:model_hyperparameters}
\end{table}

\newcommand{\thickvrule}{\vrule width 1.5pt}
\begin{table}
\begin{tabular}{|l|l!{\thickvrule}>{\centering\arraybackslash}p{0.8cm}|>{\centering\arraybackslash}p{0.8cm}|>{\centering\arraybackslash}p{0.8cm}|>{\centering\arraybackslash}p{0.8cm}|>{\centering\arraybackslash}p{0.8cm}|>{\centering\arraybackslash}p{0.8cm}|>{\centering\arraybackslash}p{0.8cm}|>{\centering\arraybackslash}p{0.8cm}|>{\centering\arraybackslash}p{0.8cm}|}
\cline{3-11}
\multicolumn{2}{c|}{}& \multicolumn{9}{c|}{\textbf{Sparsity}} \\
\cline{3-11}
\multicolumn{2}{c|}{}&1 & 2 & 4 & 8 & 16 & 32 & 64 & 128 & 256\\
\cline{3-11} 
 \specialrule{0pt}{6pt}{0pt} 
\cline{3-11}
\multicolumn{2}{c|}{}& \multicolumn{9}{c|}{Perplexity $(\downarrow)$ for given sparsity} \\
\hline
 \multirow{2}[1]{*}{\textbf{Tiny}} & \small MoSA & 22.46 & 21.76 & 20.45 & 19.24 & 18.00 & 16.90 & \textbf{16.39} & 17.27 & 18.06  \\
\hhline{~----------}
  & \small Pure MoSA & \textbf{22.46} & 22.96 & 23.30 & 24.78 & 29.76 & - & - & - & -  \\
\hhline{===========}
 \multirow{2}[1]{*}{\textbf{Small}} & \small MoSA & 16.01 & 15.74 & 15.10 & 14.48 & 13.65 & 12.97 & \textbf{12.85} &- &-   \\
\hhline{~----------}
  & \small Pure MoSA & \textbf{16.01} & 16.35& 17.16 & 19.61 & 25.41 & - & - & - & -  \\
\hline \hline 
\multirow{2}[1]{*}{\textbf{Med.}} &  \small MoSA & 13.95 & 13.52 & 12.81 & 12.16 & 11.47 & \textbf{11.06} & - & - & -   \\
\hhline{~----------}
  & \small Pure MoSA & \textbf{13.95} & 14.03 & 14.40 & 15.87& 20.63 & - & - & - & -  \\
\hhline{===========}
 \multirow{2}[1]{*}{\textbf{Large}} & \small MoSA &  12.20 & 11.33 & \textbf{10.58} & - & - & - & - & - & -  \\
\hhline{~----------}
  & \small Pure MoSA & 12.20 & \textbf{11.83} & 11.97 & - & - & - & - & - & -  \\
\hline
\end{tabular}
\newline\newline\newline
\begin{tabular}{|l|l!{\thickvrule}>{\centering\arraybackslash}p{0.8cm}|>{\centering\arraybackslash}p{0.8cm}|>{\centering\arraybackslash}p{0.8cm}|>{\centering\arraybackslash}p{0.8cm}|>{\centering\arraybackslash}p{0.8cm}|>{\centering\arraybackslash}p{0.8cm}|>{\centering\arraybackslash}p{0.8cm}|>{\centering\arraybackslash}p{0.8cm}|>{\centering\arraybackslash}p{0.8cm}|}

\cline{3-11}
\multicolumn{2}{c|}{}& \multicolumn{9}{c|}{Number of parameters for given sparsity} \\
\hline
 \multirow{2}[1]{*}{\textbf{Tiny}} & \small MoSA & 28M & 34M & 48M & 78M & 136M & 242M & 423M & 693M & 1B  \\
\hhline{~----------}
  & \small Pure MoSA & 28M & 39M & 65M & 119M & 222M & - & - & - & -  \\
\hhline{===========}
 \multirow{2}[1]{*}{\textbf{Small}} & \small MoSA & 113M & 127M & 163M & 229M & 360M & 599M &1B &- &-   \\
\hhline{~----------}
  & \small Pure MoSA & 113M & 142M & 203M & 324M & 559M & - & - & - & -  \\
\hline \hline
\multirow{2}[1]{*}{\textbf{Med.}} &  \small MoSA & 210M & 239M & 310M & 442M & 703M & 1.2B & - & - & -   \\
\hhline{~----------}
  & \small Pure MoSA & 210M & 267M & 390M & 632M& 1.1B & - & - & - & -  \\
\hhline{===========}
 \multirow{2}[1]{*}{\textbf{Large}} & \small MoSA &  516M & 650M & 943M & - & - & - & - & - & -  \\
\hhline{~----------}
  & \small Pure MoSA & 516M & 703M & 1B & - & - & - & - & - & -  \\
\hline
\end{tabular}
\newline\newline\newline
\begin{tabular}{|l|l!{\thickvrule}>{\centering\arraybackslash}p{0.8cm}|>{\centering\arraybackslash}p{0.8cm}|>{\centering\arraybackslash}p{0.8cm}|>{\centering\arraybackslash}p{0.8cm}|>{\centering\arraybackslash}p{0.8cm}|>{\centering\arraybackslash}p{0.8cm}|>{\centering\arraybackslash}p{0.8cm}|>{\centering\arraybackslash}p{0.8cm}|>{\centering\arraybackslash}p{0.8cm}|}

\cline{3-11}
\multicolumn{2}{c|}{}& \multicolumn{9}{c|}{Number of MoSA heads for given sparsity} \\
\hline
 \multirow{2}[1]{*}{\textbf{Tiny}} & \small MoSA & 0 & 13 & 31 & 69 & 142 & 276 & 505 & 848 & 1277  \\
\hhline{~----------}
  & \small Pure MoSA & 0 & 23 & 56 & 124 & 255 & - & - & - & -  \\
\hhline{===========}
 \multirow{2}[1]{*}{\textbf{Small}} & \small MoSA & 0 & 11 & 26 & 54 & 109 & 210 &381 &- &-   \\
\hhline{~----------}
  & \small Pure MoSA & 0 & 21 & 47 & 98 & 197 & - & - & - & -  \\
\hline \hline
\multirow{2}[1]{*}{\textbf{Med.}} &  \small MoSA & 0 & 11 & 26 & 54 & 109 & 210 & - & - & -   \\
\hhline{~----------}
  & \small Pure MoSA & 0 & 21 & 47& 98& 197 & - & - & - & -  \\
\hhline{===========}
 \multirow{2}[1]{*}{\textbf{Large}} & \small MoSA &  0 & 27 & 60 & - & - & - & - & - & -  \\
\hhline{~----------}
  & \small Pure MoSA & 0 & 37 & 80 & - & - & - & - & - & -  \\
\hline
\end{tabular}
\caption{Detailed statistics of the main IsoFLOP experiments from Sec.~\ref{sec:isoflop}. Models Tiny, Small, Medium, and Large are as described in App.\ref{app:models_details}. Sparsity 1 corresponds to dense baselines. Pure MoSA models for sparsities $\geq 1$ have only MoSA heads, calculated as the biggest number of heads that will not increase the FLOP budget of the dense baseline (other hyperparameters stay the same as in the baseline). MoSA models have 4 dense heads and the rest of the heads are sparse, calculated such that the flop cost of both dense and sparse heads is lower than the baseline. Therefore, the total number of heads in hybrid models (with sparsity $\geq 1$) is the number shown in the bottom table + 4. For perplexity, the best result for each row is bold. }
\label{app:detailed_isoflop_results}
\end{table}

\end{document}